\documentclass{article}

\usepackage{microtype}
\usepackage{graphicx}
\usepackage{subfigure}
\usepackage{booktabs} % for professional tables
\usepackage{multirow} % For \multirow
\usepackage{multicol}
\usepackage{subcaption}
\usepackage{xcolor}
\usepackage{tabularx}
\usepackage{adjustbox}
\usepackage{color, colortbl}
\definecolor{Gray}{gray}{0.9}

\usepackage{pifont}% http://ctan.org/pkg/pifont
\newcommand{\cmark}{\ding{51}}%
\newcommand{\xmark}{\ding{55}}%

\usepackage{bbm}
\usepackage{bm}

\usepackage{hyperref}

\usepackage[accepted]{icml2024}

\usepackage{amsmath}
\usepackage{amssymb}
\usepackage{mathtools}
\usepackage{amsthm}

\usepackage[capitalize,noabbrev]{cleveref}

\def\ie{\emph{i.e.}}
\def\eg{\emph{e.g.}}

\theoremstyle{plain}
\newtheorem{theorem}{Theorem}[section]
\newtheorem{proposition}[theorem]{Proposition}

\theoremstyle{definition}

\theoremstyle{remark}

\usepackage[textsize=tiny]{todonotes}

\icmltitlerunning{Reducing Fine-Tuning Memory Overhead by Approximate and Memory-Sharing Backpropagation}

\begin{document}

\twocolumn[
\icmltitle{Reducing Fine-Tuning Memory Overhead by \\
Approximate and Memory-Sharing Backpropagation
}



\icmlsetsymbol{equal}{*}
\begin{icmlauthorlist}
\icmlauthor{Yuchen Yang}{nankai}
\icmlauthor{Yingdong Shi}{shang}
\icmlauthor{Cheems Wang}{sch1}
\icmlauthor{Xiantong Zhen}{sch2}
\icmlauthor{Yuxuan Shi}{nankai}
\icmlauthor{Jun Xu}{nankai}
\end{icmlauthorlist}
\icmlaffiliation{nankai}{School of Statistics and Data Science, Nankai University, Tianjin, China}
\icmlaffiliation{shang}{School of Information Science and Technology, ShanghaiTech University, Shanghai, China}
\icmlaffiliation{sch1}{Department of Automation, Tsinghua University, Peking, China}
\icmlaffiliation{sch2}{Central Research Institute, United Imaging Healthcare, Co., Ltd.}

\icmlcorrespondingauthor{Jun Xu}{nankaimathxujun@gmail.com}

\icmlkeywords{Machine Learning, ICML, Fine-tuning, Memory Reducing, Efficient Training, BP}

\vskip 0.3in
]



\printAffiliationsAndNotice{}  % leave blank if no need to mention equal contribution

\begin{abstract}
Fine-tuning pretrained large models to downstream tasks is an important problem, which however suffers from huge memory overhead due to large-scale parameters.
This work strives to reduce memory overhead in fine-tuning from perspectives of activation function and layer normalization.
To this end, we propose the Approximate Backpropagation (Approx-BP) theory, which provides the theoretical feasibility of decoupling the forward and backward passes.
We apply our Approx-BP theory to backpropagation training and derive memory-efficient alternatives of GELU and SiLU activation functions, which use derivative functions of ReLUs in the backward pass while keeping their forward pass unchanged.\ In addition, we introduce a Memory-Sharing Backpropagation strategy, which enables the activation memory to be shared by two adjacent layers, thereby removing activation memory usage redundancy.
Our method neither induces extra computation nor reduces training efficiency.
We conduct extensive experiments with pretrained vision and language models, and the results demonstrate that our proposal can reduce up to $\sim$$30\%$ of the peak memory usage.
Our code is released at \href{https://github.com/yyyyychen/LowMemoryBP}{github}.
\end{abstract}
\begin{figure}[h!]
% \vskip -0.2in
\begin{center}
\includegraphics[width=\linewidth]{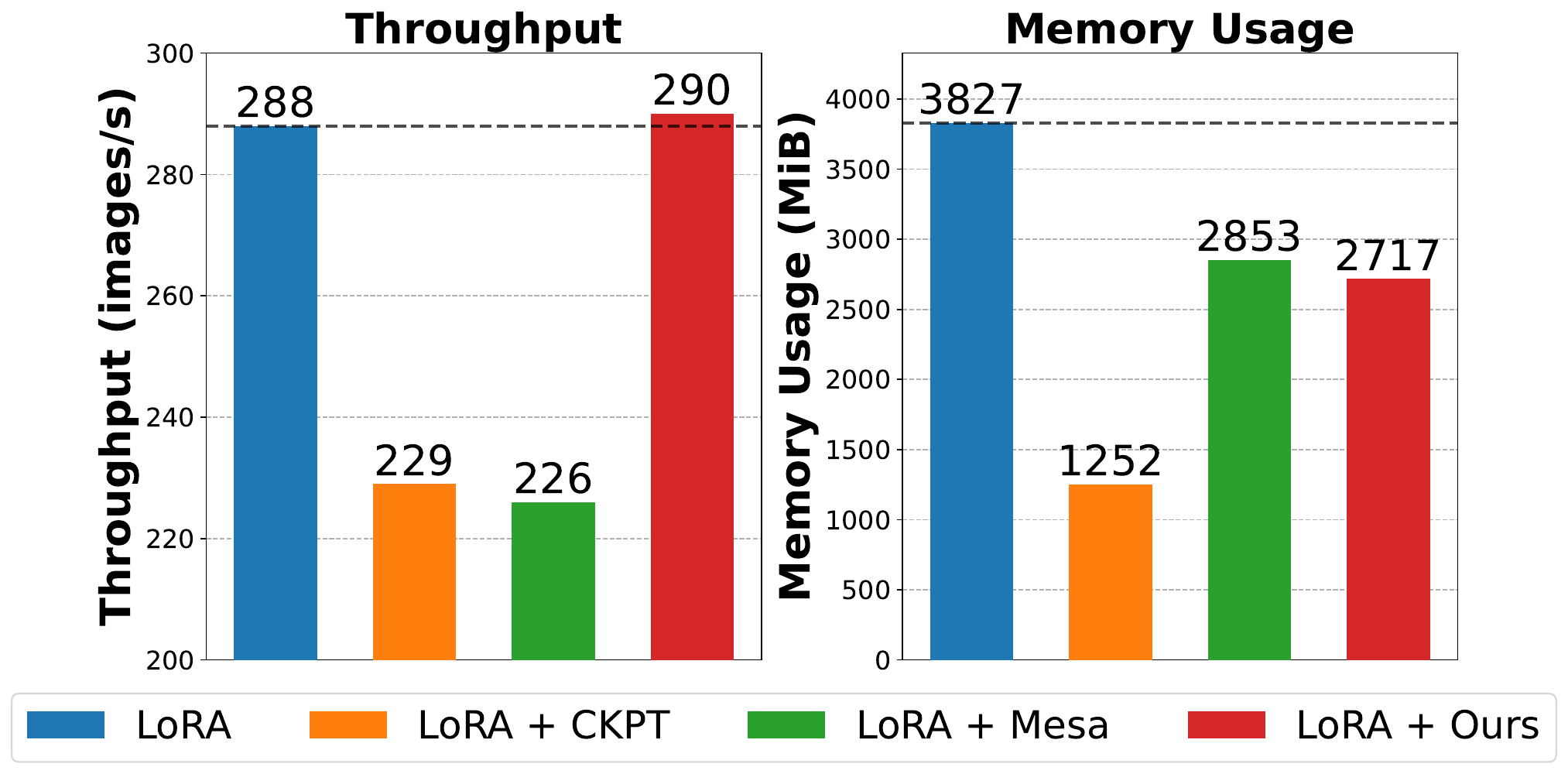}
\vskip -0.15in
\caption{\textbf{Throughput (images/s) and memory usage (MiB) with LoRA~\cite{hu2022lora} (rank $=4$, batch size $=64$) on fine-tuning pretrained ViT-B~\cite{dosovitskiy2020vit} with CIFAR10/100~\cite{krizhevsky2009learning} and FGVC~\cite{jia2022vpt}}.
``LoRA + CKPT'': LoRA with gradient-checkpointing~\cite{chen2016training} on every block.
``LoRA + Mesa'': LoRA with 8-bit activation quantization on GELU and LayerNorm~\cite{pan2021mesa}.
``LoRA + Ours'': LoRA with our ReGELU2 and MS-LN.
More details are provided in \cref{sec:experiments}.
}
\label{fig:banner}
\end{center}
\vskip -0.2in
\end{figure}

\section{Introduction}
\label{introduction}

Ever since the emergence of large models like GPTs~\cite{radford2019language}, how to fine-tune them efficiently on downstream tasks has become an important problem~\cite{hu2022lora}.
However, the unaffordable activation memory overhead largely limits their applications to memory-constrained hardware like edge devices.
For this, it is essential to investigate memory reduction strategies for parameter-efficient fine-tuning (PEFT).
A common strategy of PEFT~\cite{houlsby2019parameter,liu2021p,hu2022lora,jia2022vpt} is parameter freezing, which mainly reduces activation memory usage brought by linear projection layers.
However, the activation memory overhead from non-linear modules in transformers still occupies a large part of the total usage, \eg,~$\sim 63\%$ in ViT~\cite{dosovitskiy2020vit} and $\sim 74\%$ in LLaMA~\cite{touvron2023llama} (\cref{fig:activation_memory_ratio}).

There are three main non-linear modules in a typical transformer: self-attention, activation function, and layer normalization.
Previous efforts have been mainly devoted to reducing the memory complexity of vanilla self-attention~\cite{child2019generating,kitaev2020reformer,beltagy2020longformer,dao2022flashattention}.
Among them, FlashAttention \cite{dao2022flashattention} is a highly optimized implementation with linear memory complexity.
Hereafter, transformers put a large part of activation memory usage in activation function and layer normalization.
Nevertheless, these two modules draw little attention on activation memory reduction, though they are widely used in transformers~\cite{liu2019roberta,touvron2022deit} and others~\cite{tolstikhin2021mixer,yu2022metaformer}. 
% (\cref{fig:activation_memory_ratio}).

%
Non-linear activation functions like GELU~\cite{hendrycks2023gaussian} and SiLU~\cite{hendrycks2023gaussian,elfwing2017sigmoidweighted,ramachandran2017searching} need the whole input tensor to compute the gradients in regular backpropagation (BP), and suffer from huge activation memory usage.
% significant
To avoid performance degradation, one may prefer to initialize the large model with the pretrained weights before fine-tuning it.
To this end, it is safe to avoid changing the forward pass of activation functions.
A natural yet crucial question arises: \textit{\textbf{is it possible to reduce the activation memory usage by only changing the backward pass?}} 

This paper provides positive feedback to the above question by developing an approximate backward pass as an alternative to the exact BP process.
To achieve this goal, we propose safely decoupling the forward and backward passes with a new Approximate BackPropagation (Approx-BP) theory.
Our Approx-BP theory reveals that if primitive functions are close in functional space, then derivatives can be substituted for each other in the training.
Based on our Approx-BP theory, the pretrained models using a highly non-linear activation function could replace their non-linear derivatives with a moderately linear derivative that requires less activation memory.
We apply this theory to GELU and SiLU, and derive our ReGELU2 and ReSiLU2 in which the activation memory usage is only 2 bits per element.

As for layer normalization~\cite{ba2016layer}, we observe redundancy in the activation memory within it and the subsequent linear layers. To avoid this redundancy, we introduce a Memory-Sharing BP (MS-BP) strategy and establish a sufficient condition under which a layer can share its activation memory with the following layer.
By merging the affine parameters of LayerNorm and RMSNorm~\cite{zhang2019root} into the following linear layers with an adapted derivative calculation manner, we propose memory-sharing LayerNorm (MS-LN) and RMSNorm (MS-RMSNorm) to satisfy the condition of our MS-BP strategy and share activation memory usage with the following linear layers.

Without any extra computation cost, our method will not affect the training throughput of full fine-tuning or PEFT methods like LoRA while further reducing their activation memory usage (\cref{fig:banner}).
Experiments on ViT \cite{dosovitskiy2020vit} and LLaMA \cite{touvron2023llama} show that our method can reduce their peak GPU memory usage in fine-tuning by $\sim$$ 30\%$, with comparable performance to those by full fine-tuning, LoRA \cite{hu2022lora}, LoRA-FA \cite{zhang2023lorafa}, or QLoRA \cite{dettmers2023qlora}.

In summary, the contributions of this work are three-fold:
\vspace{-2mm}
\begin{itemize}
\vspace{-2mm}
\item We propose the Approximate Backpropagation (Appro- 
x-BP) theory, which supports the feasibility of decoupling the forward and backward passes in backpropagation training.
Under our Approx-BP, we derive our ReGELU2 and ReSiLU2 as alternatives of GELU and SiLU, respectively, to share their primitives while possessing a 2-bit step function as the derivative.
\vspace{-1.1mm}
\item We provide a Memory-Sharing BP (MS-BP) strategy and apply it to layer normalization.
The resulting MS-LN and MS-RMSNorm remove the redundant
activation memory with the following linear layers.
\vspace{-1.1mm}
\item Our method has no extra computational cost and does not affect the training throughput or the fine-tuning networks' inference accuracy.
\end{itemize}

\section{Related Work}
\label{related works}
Here, we briefly introduce the related research on reducing the activation memory usage in network training.

% \subsection{Parallel Training}
% %
% Parallel training can be roughly classified as Data Parallel (DP) and Model Parallel (MP).
% %
% In Distributed Parallelism (DP), each computing node independently performs a full backpropagation process and intermittently exchanges gradients, and in some cases parameters \cite{rajbhandari2020zero}, with other nodes.
% %
% In Model Parallel (MP), each computing node performs a partial forward pass and backward pass and communicate the intermediate activation tensors with other nodes \cite{huang2019gpipe,narayanan2021efficient}.
% %
% The reason for memory efficiency in parallel training is that these methods systematically place data, parameters and intermediate activation tensors on multiple GPUs or CPU \cite{ren2021zero}, so that the memory usage in the training, including activation memory, can be split to multiple devices.
% %
% Although parallel training can reduce the activation memory overhead per GPU, it will not reduce the total activation memory and will bring additional communication overhead.
% %
\subsection{Activation Recomputation}
The activation recomputation \cite{chen2016training} (also called gradient checkpointing) avoids saving the intermediate activation in the forward pass of a network layer by recomputing it in the backward pass.
% quadratic-complexity
It is widely used on self-attention~\cite{korthikanti2023reducing} to reduce the activation memory usage but at the cost of extra computation.

Later, FlashAttention \cite{dao2022flashattention} optimizes the complexity of activation recomputation in self-attention, which is implemented by an efficient CUDA kernel.
Due to preserving the training process while effectively reducing the activation memory, gradient checkpointing is widely used in fine-tuning large models with GPU constraints.
However, it suffers from a remarkable side affect of additional training duration, \eg, $\sim$$20\%$ in LoRA fine-tuning (\cref{fig:banner}) when used at every block of the fine-tuning network.

Our method also changes the regular BP process, but avoids recomputation to preserve the training efficiency.

\subsection{Activation Quantization}
Network training in mixed precision \cite{micikevicius2017mixed} is feasible to execute most of computations in half-precision floats (16-bit) in forward and backward passes.
Besides reducing memory usage, this can also accelerate the training speed since half-precision computation is supported inherently in modern GPUs.
8-bit training is allowed in CNNs with tolerant performance loss~\cite{banner2018scalable}.

To avoid global quantization in both forward and backward passes, activation compression training (ACT) \cite{chakrabarti2019backprop} executes the forward pass in the originally high precision, then stores activation tensors by low precision quantization, and finally dequantizes these tensors back to the original precision in the backward pass.
Later, ActNN \cite{chen2021actnn} stores activation tensors in 2-bit precision for training CNNs, greatly reducing activation memory usage by $\sim$$12\times$.
Mesa \cite{pan2021mesa} uses a customized 8-bit activation quantization strategy for training transformers.
AC-GC \cite{evans2021ac} established a direct relationship between quantization error and training convergence by automatically selecting the compression ratios. GACT \cite{liu2022gact} introduced an adaptive compression strategy for general network architectures, which utilizes the empirical variance of the gradients to estimate the sensitivity of quantized activation tensors. ALAM \cite{woo2023alam} quantizes the group mean estimator and calculates the sensitivity by the empirical variance of the gradients' norm to allocate adaptive compression bits.
When applied to transformers, these ACT methods generally reduce more activation memory usage than gradient checkpointing.
However, frequent quantization and dequantization in training adversely affect the training throughput of transformers~\cite{wang2023division}.

Our method avoids quantization and dequantization during training, and thus keeps the training throughput.

\subsection{Parameter-Efficient Fine-Tuning}

Parameter-Efficient Fine-Tuning (PEFT) is widely used for transformers due to little memory usage in storing the gradients of trainable parameters or the optimizer states, \eg, AdamW \cite{loshchilov2017decoupled}.
Adapter \cite{houlsby2019parameter} inserts a two-layer MLP with residual connection after each FFN block.
Later, BitFit \cite{zaken2021bitfit} only fine-tunes the bias and freezes other parameters in the transformers.
Prompt Tuning is also studied in \cite{lester2021power,li2021prefix,liu2021p,jia2022vpt} to prepend extra learnable prompt tokens in self-attention.
Recently, LoRA \cite{hu2022lora} and its variants \cite{zhang2023adaptive,zhang2023adaptive,jie2023fact,zhang2023lorafa,kopiczko2023vera} are widely used for scalable fine-tuning power with no extra inference overhead. These methods mainly use low-rank matrices to fine-tune linear layers.
By freezing ``LoRA-A'' parameters, the variant LoRA-FA \cite{zhang2023lorafa} can eliminate most of the activation memory costs from linear layers in fine-tuning.

Though using few trainable parameters, these PEFT methods still consume the same order of magnitude of activation memory usage as those used in full fine-tuning.
An exception LST~\cite{sung2022lst} uses a ladder side model to avoid backward passes through the pretrained modules. However, it performs inferior to LoRA and brings extra memory overhead and latency in the inference stage.

Unlike LoRAs, our work aims to reduce the activation memory usage from non-linear layers in transformers.

\subsection{Activation Approximation}

The work of AAL \cite{woo2022learning} introduced auxiliary activation to participate the backward pass instead of the original input activation in linear layers.
The auxiliary activation is typically the activation from the previous block or the sign of the original activation.
The work of \cite{jiang2022back} introduced an asymmetric sparsifying strategy to obtain sparse activation features for back-propagation, while keeping dense forward activation features.
These two works both showed the compatibility with Mesa \cite{pan2021mesa} in their papers.
Since our method is functionally similar to Mesa, they are also compatible with our method.

\section{Preliminary}
\label{preliminary}
% We introduce the basics of fine-tuning and analyze its activation memory usage.
%
\subsection{Fine-Tuning}
\label{pre:fine-tuning}
Denote $\bm{x}\in\mathbb{R}^{p_0}$ as the network input vector under the data distribution $\mathcal{D}$, \ie, $\bm{x}\sim\mathcal{D}$.
For an $L$-layer neural network $\bm{f}(\bm{x},\bm{\theta})$, the output feature vector of $i$-th hidden layer $\bm{h}^i$ is denoted as $\bm{z}^{i} = \bm{h}^i(\bm{z}^{i-1}, \bm{\theta}^i) = \bm{h}_{\bm{\theta}}^i(\bm{z}^{i-1}) \in \mathbb{R}^{p_i}$, where $\bm{z}^0=\bm{x}$, $\bm{\theta} = [{\bm{\theta}^1}^\top, ..., {\bm{\theta}^L}^\top]^\top \in \mathbb{R}^M$ and $\bm{\theta}^i$ is the straightened vector of network parameters of the $i$-th hidden layer $\bm{h}^i$.
The network can be formulated as
\vspace{-1mm}
\begin{equation}
\bm{z}^L=\bm{f}(\bm{x}, \bm{\theta}) = \bm{h}_{\bm{\theta}}^L \circ \bm{h}_{\bm{\theta}}^{L-1} \circ ... \circ \bm{h}_{\bm{\theta}}^{1}(\bm{x}),
\label{eq:netword}
\vspace{-1mm}
\end{equation}
where ``$\circ$'' denotes layer composition. For simplicity, we define the set of all feature vectors by $\bm{z} = [{\bm{z}^1}^\top, ..., {\bm{z}^L}^\top]^\top$ and express the backward pass of network training as:
\begin{equation}
\bm{g}
\triangleq
\bm{g}(\ell(\bm{z}^L), \bm{z}, \bm{\theta}) = \nabla_{\bm{\theta}} \ell(\bm{z}^L),
\label{eq:backward_pass}
% \vspace{-1.1mm}
\end{equation}
where $\ell(\bm{z}^L) = \ell(\bm{f}(\bm{x},\bm{\theta}))$ is the loss function, $\bm{g}$ is a composite function of the derivatives of $\ell$ and $\bm{h}$, \ie, $\mathrm{d}\ell$ and $\{\mathrm{d} \bm{h}^i\}_{i=1}^{L}$, respectively.
Then the parameter update at $t$-th iteration in regular BP can be expressed as:
\begin{equation}
\bm{\theta}_{t+1} = \bm{\theta}_{t} - \eta \bm{g}_t = \bm{\theta}_{t} - \eta \nabla_{\bm{\theta}_t} \ell(\bm{z}^L_t).
\label{eq:backpropagation}
% \vspace{-1.1mm}
\end{equation}
The main characteristic of fine-tuning is that the model parameters are initialized as the pretrained weights, \ie, $\bm{\theta}_0=\bm{\theta}_{\rm pretrained}$.
Different from the ``training from scratch'', the initial model $\bm{f}(\bm{x}, \bm{\theta}_0)$ to be fine-tuned usually already has potential capability on the downstream tasks.
\subsection{Activation Memory Usage in Fine-Tuning}
In general, all intermediate feature vectors $\{\bm{z}^i\}_{i=1}^L$ may participate in calculating gradients in the backward pass \eqref{eq:backward_pass}.
However, according to the specific layers in the network, we do not need to store all $\{\bm{z}^i\}_{i=1}^L$ into activation memory in practice.
For example, freezing partial parameters is a widely used fine-tuning technique by PEFT methods.
A frozen linear layer can be expressed as:
\begin{equation}
% \vspace{-1.1mm}
\bm{z}^i = \bm{h}^i(\bm{z}^{i-1}) = \bm{W}_{\rm frozen} \bm{z}^{i-1} + \bm{b}_{\rm frozen},
\label{eq:frozen_linear}
% \vspace{-1.1mm}
\end{equation}
where the frozen weight and bias need no gradient, avoiding storing the input feature $\bm{z}^{i-1}$ into activation memory.
As a representative PEFT method, LoRA~\cite{hu2022lora} is briefly analyzed here and we express its adapting layer as
\vspace{-1.1mm}
\begin{equation}
\label{eq:lora}
\bm{z}^i = \bm{h}^i(\bm{z}^{i-1}) = \bm{W}_{\rm frozen} \bm{z}^{i-1} + 
\bm{B}\bm{A}\bm{z}^{i-1} + \bm{b}_{\rm frozen},
\vspace{-1.1mm}
\end{equation}
where $\bm{A} \in \mathbb{R}^{r \times p_{i-1}}$ and $\bm{B} \in \mathbb{R}^{p_{i} \times r}$ are trainable parameters.
The stored features in activation memory are $\bm{z}^{i-1} \in \mathbb{R}^{p_{i-1}}$ and $\bm{Az}^{i-1} \in \mathbb{R}^r$.
Since $r \ll p_{in}$, the activation memory usage in a LoRA adapting layer is only slightly larger than that storing $\bm{z}^{i-1}$ in the linear layer.
Besides, LoRA-FA~\cite{zhang2023lorafa} further freezes the projection matrix $\bm{A}$ in the LoRA adapting layers~\eqref{eq:lora}, and only stores the $r$-dimensional $\bm{Az}^{i-1}$ in activation memory.

\begin{figure}[t]
\begin{center}
\centerline{\includegraphics[width=\columnwidth]{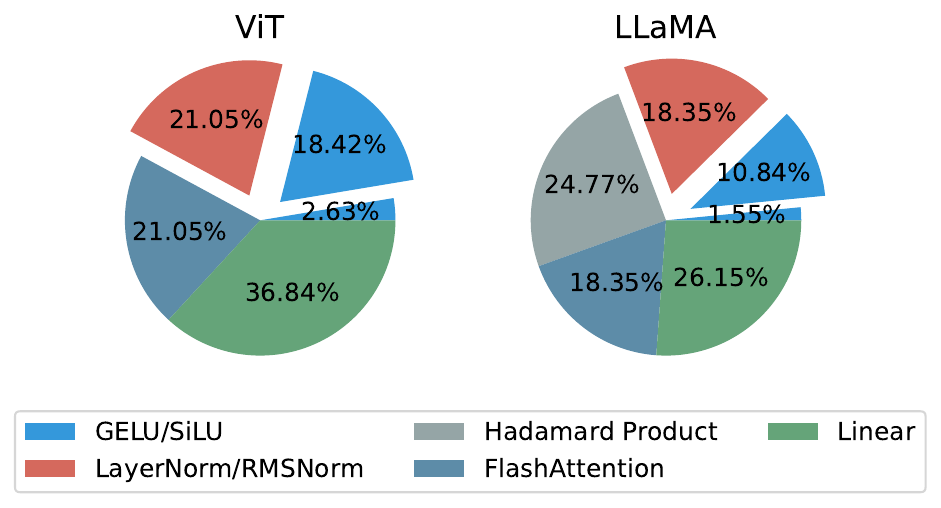}}
\vskip -0.2in
\caption{\textbf{Composition of activation memory usage in ViT and LLaMA}. For LLaMA, we use LLaMA-13B as an example. Our method is feasible to reduce the activation memory usage of GELU/SiLU and LayerNorm/RMSNorm (the split parts).}
\label{fig:activation_memory_ratio}
\end{center}
\vskip -0.2in
\end{figure}

Although freezing techniques can reduce the activation memory usage in linear layers (including LoRA adapting layers), the activation memory overhead in non-linear layers is still expensive.
Among these non-linear layers, activation function and layer normalization bring a large part of activation memory usage.
In \cref{fig:activation_memory_ratio}, we illustrate the memory usage ratios of different modules in ViT~\cite{dosovitskiy2020vit} and LLaMA~\cite{touvron2023llama}. One can see that in ViT both GELU and LayerNorm occupy $21.05\%$ of the total activation memory usage, while in LLaMA $12.39\%$ and $18.35\%$ memory usage are from SiLU and RMSNorm, respectively (please refer to \cref{appendix:analysis_on_block} for more details).

\section{Approximate Backpropagation}
\label{sec:approximate_bp}
A large model like LLaMA exhibits strong representation capability with its pretrained weights, which are usually more crucial than the fine-tuning itself to its performance on downstream tasks.
Therefore, it is reasonable to fine-tune the large model with its architecture the same as the original design (see \cref{appendix:possibility_of_sf} for our empirical investigation).

To provide flexible fine-tuning scheme, in this section, we show the possibility of substituting the backward pass while remaining the forward pass of the pretrained model.
In \cref{subsec:approx-bp_theory}, we present our Approximate Backpropagation (Approx-BP) theory to demonstrate the theoretical feasibility of decoupling the forward and backward passes.
In \cref{method:regelu2}, under the guidance of our Approx-BP theory, we derive ReGELU2 and ReSiLU2 as memory-efficient alternatives of GELU and SiLU, respectively in transformers.

\subsection{Approx-BP Theory}
\label{subsec:approx-bp_theory}
We introduce an approximate network $\widetilde{\bm{f}}$ that shares the same parameters $\bm{\theta}$ with $\bm{f}$ in Eqn.~\eqref{eq:netword}, \ie,
\vspace{-1mm}
\begin{equation}
\label{eq:approx_netword}
\widetilde{\bm{f}}(\bm{x}, \bm{\theta}) = \widetilde{\bm{h}}_{\bm{\theta}}^L \circ \widetilde{\bm{h}}_{\bm{\theta}}^{L-1} \circ ... \circ \widetilde{\bm{h}}_{\bm{\theta}}^{1}(\bm{x}).
\vspace{-1mm}
\end{equation}
The loss function of $\widetilde{\bm{f}}$ is similarly denoted as $\ell(\widetilde{\bm{z}}^L) = \ell(\widetilde{\bm{f}}(\bm{x}, \bm{\theta}))$, and its backward pass is denoted as
\vspace{-1mm}
\begin{equation}
\widetilde{\bm{g}}
\triangleq
\widetilde{\bm{g}}(\ell(\widetilde{\bm{z}}^L), \widetilde{\bm{z}}, \bm{\theta}) = \nabla_{\bm{\theta}} \ell(\widetilde{\bm{z}}^L).
\vspace{-1mm}
\end{equation}
Here, the definitions of $\widetilde{\bm{z}}$ and $\widetilde{\bm{g}}$ are the counterparts of $\bm{z}$ and $\bm{g}$ in \cref{pre:fine-tuning}, respectively.
In order to approximate the backward pass of network training in Eqn.~\eqref{eq:backward_pass}, we formulate our Approx-BP as
\vspace{-1mm}
\begin{equation}
\widehat{\bm{g}}
\triangleq
\widetilde{\bm{g}}(\ell(\bm{z}^L), \bm{z}, \bm{\theta})
\approx
\bm{g}(\ell(\bm{z}^L), \bm{z}, \bm{\theta}).
\vspace{-1mm}
\end{equation}
Then we replace the gradient update in regular BP \eqref{eq:backpropagation} by
\vspace{-1mm}
\begin{equation}
\bm{\theta}_{t+1} = \bm{\theta}_{t} - \eta \widehat{\bm{g}}_t.
\vspace{-1mm}
\label{eq:abp_update}
\end{equation}
By decoupling the forward and backward passes, our Approx-BP is feasible to flexibly fine-tune large models.

By Triangle Inequality, we can derive an insightful property about our Approx-BP as follows:
\vspace{-1mm}
\begin{equation}
\begin{aligned}
\Vert \widehat{\bm{g}} - \bm{g} \Vert
\leq
&\Vert \widehat{\bm{g}} - \widetilde{\bm{g}} \Vert + \Vert \widetilde{\bm{g}} - \bm{g} \Vert \\
=
&\Vert \widetilde{\bm{g}}(\ell(\bm{z}^L), \bm{z}, \bm{\theta}) - \widetilde{\bm{g}}(\ell(\widetilde{\bm{z}}^L), \widetilde{\bm{z}}, \bm{\theta}) \Vert 
\ +
\\
&\Vert \nabla_{\bm{\theta}} \ell(\widetilde{\bm{z}}^L) - \nabla_{\bm{\theta}} \ell(\bm{z}^L) \Vert.
\label{gradient_bound}
\end{aligned}
\vspace{-1mm}
\end{equation}
The inequality \eqref{gradient_bound} indicates that approximate BP $\widehat{\bm{g}}$ and the regular BP $\bm{g}$ differs in the intermediate outputs of forward pass $\Vert \bm{z} - \widetilde{\bm{z}} \Vert$, if functions $\widetilde{\bm{g}}$ and $\ell$ are in proper continuity.
This observation motivates us to design proper alternatives to replace the derivatives of (non-linear) modules in a neural network, as long as their primitive functions are close enough in the functional space.
We describe the degree of approximation in our Approx-BP by the following theorem.
\begin{theorem}
\label{thm:abp}
Under the definitions in \cref{subsec:approx-bp_theory}, assume that:\\
\textbf{A1.}~$\widetilde{\bm{g}}(\ell(\bm{z}^L), \bm{z}, \bm{\theta})$ is uniformly Lipschitz continuous w.r.t.~$\ell(\bm{z}^L)$ and $\bm{z}$.\\
\textbf{A2.}~$\ell(\bm{z}^L)$ is Lipschitz continuous. $\bm{h}^i(\bm{z}^{i-1},\bm{\theta}^i)$ is uniformly Lipschitz continuous w.r.t.~$\bm{z}^{i-1}$ for $i=2,...,L$.\\
\textbf{A3.}~$\ell(\bm{f}(\bm{x}, \bm{\theta}))$ and $\ell(\widetilde{\bm{f}}(\bm{x}, \bm{\theta}))$ are twice differentiable w.r.t.~$\bm{\theta}$ with uniformly bounded induced norm of their Hessian matrices.
Then, $\exists\ \alpha > 0$, $\forall\ \bm{x}, \bm{\theta}$, we have
\vspace{-1mm}
\begin{equation}
\begin{aligned}
&\Vert \widehat{\bm{g}} - \bm{g} \Vert_2 \\
&\leq \alpha \Big(\sum^{L}_{i=1} \sup_{\bm{z}^{i-1}, \bm{\theta}^i} \Vert \bm{h}^i(\bm{z}^{i-1}, \bm{\theta}^{i}) - \widetilde{\bm{h}}^i(\bm{z}^{i-1}, \bm{\theta}^{i}) \Vert_2 +  \\
&\sqrt{\sum^{L}_{i=1} \sup_{\bm{z}^{i-1}, \bm{\theta}^i} \Vert \bm{h}^i(\bm{z}^{i-1}, \bm{\theta}^{i}) - \widetilde{\bm{h}}^i(\bm{z}^{i-1}, \bm{\theta}^{i}) \Vert_2} \ \Big).
\end{aligned}
\end{equation}
\vspace{-1mm}
\end{theorem}
\vspace{-1mm}
%
% Note that the assumptions in \cref{thm:abp} are loose enough to satisfy even when the network contains operators like ReLU \cite{nair2010rectified}, whose derivatives are step functions.
%
Although the networks containing ReLUs \cite{nair2010rectified} do not strictly satisfy the assumptions in \cref{thm:abp}, the violations only happen in a zero measure set.
In the practical training, we can safely conceive a smoothing curve at the neighborhood of zero point in ReLU.
Next, we demonstrate the convergence of our Approx-BP theory by another theorem described as follows.
\begin{theorem}
\label{thm:abpt}
Suppose data $\bm{x}$ follows the distribution $\mathcal{D}$. Denote $T$ as the total iteration number.
Assume that:\\
\textbf{A1.}~$\ell(\bm{f}(\bm{x}, \bm{\theta}))$ is continuously differentiable w.r.t. $\bm{\theta}$, and $\nabla_{\bm{\theta}} \ell(\bm{f}(\bm{x}, \bm{\theta}))$ is $\beta$-Lipschitz continuous w.r.t. $\bm{\theta}$. \\
\textbf{A2.}~$\ell(\bm{f}(\bm{x}, \bm{\theta}))$ is bounded below by a constant $\ell^{*}$. \\
\textbf{A3.}~$\exists\ \sigma > 0$, for $\forall \bm{\theta}$, $\mathbb{E}_\mathcal{D} \Vert \widehat{\bm{g}} - \bm{g} \Vert^2_2 < \sigma^2$. \\
Then, for all $\eta < \frac{1}{2\beta}$, if we run Approx-BP training defined in \eqref{eq:abp_update}, we have
\vspace{-1mm}
\begin{equation}
\begin{aligned}
\mathop{\min}_{t \in \{0, ..., T-1\}} &\mathbb{E}_\mathcal{D} \Vert \nabla_{\bm{\theta}} \ell(\bm{f}(\bm{x}, \bm{\theta}_t)) \Vert^2_2 
\\
\leq &\frac{4(\mathbb{E}_\mathcal{D} \ell(\bm{f}(\bm{x}, \bm{\theta}_0)) - \ell^*)}{\eta T} + 6\sigma^2.
\end{aligned}
\vspace{-1mm}
\end{equation}
\vspace{-1mm}
\end{theorem}
\vspace{-1mm}
From \cref{thm:abp} and \cref{thm:abpt}, we conclude that the learning capability of the network $\bm{f}(\bm{x},\bm{\theta})$ with our Approx-BP theory mainly correlates to the functional closeness between the original layers $\bm{h}$ and the approximate layers $\widetilde{\bm{h}}$.

The theoretical analysis reveals that our Approx-BP theory can work as a feasible framework to decouple the forward and backward passes, with guaranteed training convergence. In contrast, the regular BP in network training links the two opposite passes in a balanced scale of memory overhead. Instead, our Approx-BP can potentially break the scale balance, and is feasible to reduce the activation memory.

\subsection{Approx-BP on Activation Functions}
\label{method:regelu2}

Transformers \cite{radford2019language, dosovitskiy2020vit, touvron2022deit, touvron2023llama} usually use GELU or SiLU \cite{hendrycks2023gaussian} as the non-linear activation function in MLP blocks.
GELU and SiLU~\cite{hendrycks2023gaussian} usually boost the network performance against ReLU~\cite{nair2010rectified} in various vision and language tasks.
However, GELU and SiLU need to store the whole 16-bit input tensor for backward pass, while ReLU only needs to store the 1-bit signs of the input tensor elements.
% efficiency --->> reduction?
Therefore, for consideration of memory efficiency, we propose to combine multiple ReLUs to approximate the regular BP process of GELU and SiLU.
Denote $h$ as the activation function of GELU or SiLU, we have
\begin{equation*}
\vspace{-1.5mm}
\begin{aligned}
h(x) &= \mathrm{GELU}(x) = \frac{x}{2}(1 + \mathrm{erf}(\frac{x}{\sqrt{2}})) \\
\text{or} \quad h(x) &= \mathrm{SiLU}(x) = \frac{x}{1 + e^{-x}}.
\end{aligned}
\end{equation*}
We define a combination of multiple ReLUs as
\begin{equation}
\begin{aligned}
\widetilde{h}_{\bm{a},\bm{c}}(x) &= \mathop{\sum}_{i=1}^{2^k-2}a_i \mathrm{ReLU}(x - c_i)\ + \\
&(1 - \mathop{\sum}_{i=1}^{2^k-2}a_i)\mathrm{ReLU}(x - c_{2^k-1}), \\
\text{s.t.} \mathop{\sum}_{i=1}^{2^k-2} a_i c_i &+(1 - \mathop{\sum}_{i=1}^{2^k-2}a_i) c_{2^k-1} = 0,
\label{eq:relus_family}
\end{aligned}
\end{equation}
where the $i$-th element $a_i$ (or $c_i$) of $\bm{a}$ (or $\bm{c}$) indicates the weight (or bias) of the $i$-th ReLU in our combined ReLUs. Here we use $2^k-1$ ReLUs in $\widetilde{h}_{\bm{a},\bm{c}}$ and $k$ is the required bit number of activation memory for derivative calculation.

\begin{proposition}
The combination function $\widetilde{h}_{\bm{a},\bm{c}}$ of multiple ReLUs in Eqn.~\eqref{eq:relus_family} has the following two properties:
\vspace{-1.5mm}
\begin{enumerate}
\item It has the same limiting behavior with the the activation function $h(x)$, i.e.,
$\mathop{\lim}_{x \rightarrow \infty} h(x) - \widetilde{h}_{\bm{a},\bm{c}}(x) = 0$.\\
\vspace{-1.5mm}
\item Its derivative is a $2^k$-segment step function that need k bits of activation memory for derivative calculation.
\end{enumerate}  
\vspace{-1.5mm}
\end{proposition}

\begin{figure}[t]
\begin{center}
\centerline{\includegraphics[width=\columnwidth]{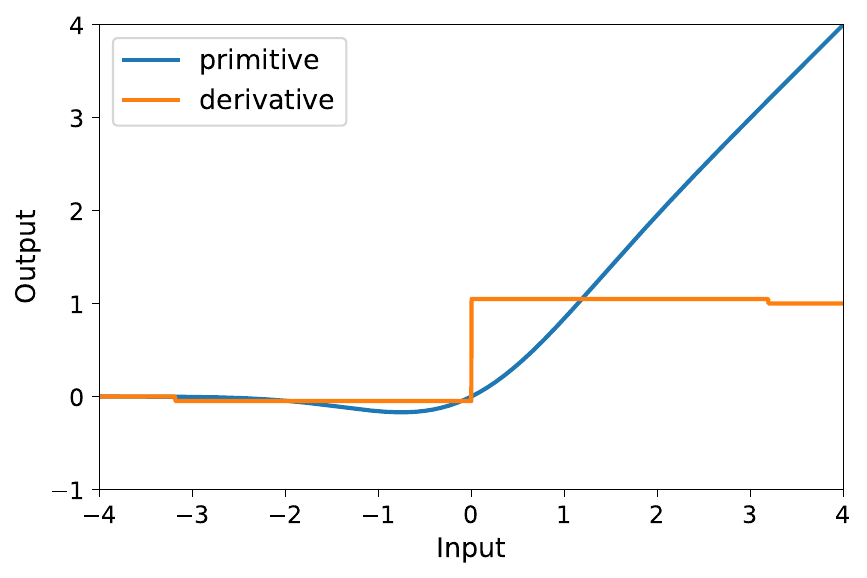}}
\vskip -0.2in
\caption{\textbf{Plot of our ReGELU2}. The primitive function is still GELU, while the derivative function is a 4-segment step function that need 2 bits of activation memory for derivative calculation.}
\label{fig:regelu2}
\end{center}
\vskip -0.3in
\end{figure}

Here, we set $k = 2$ to reduce the activation memory usage in activation functions.
According to our Approx-BP theory, we should set the parameters in \eqref{eq:relus_family}, so that $\widetilde{h}_{\bm{a},\bm{c}}(x)$ could be close to $h(x)$ in the function space.
% According to our Approx-BP theory, the parameters in our combined ReLUs $\widetilde{h}_{\bm{a},\bm{c}}(x)$ are obtained by solving the following optimization objective function:
% \vspace{-2mm}
% \begin{equation}
% \label{eqn:optimization}
% \vspace{-2mm}
% \mathop{\mathrm{min}}_{\bm{a},\bm{c}} \mathbb{E}_x\Vert h(x) - \widetilde{h}_{\bm{a},\bm{c}}(x) \Vert^2_2.
% \end{equation}
% By assuming that the input $x$ of $h$ are uniformly distributed over the define domain of $h$,
% % solvable ---> feasible
% we instead resort to solving a numerically feasible version of the problem \eqref{eqn:optimization}:
To implicitly fulfill the constraint in \eqref{eq:relus_family} and put uniform importance to the define domain, we solve the following feasible problem:
\begin{equation}
\label{eq:objective}
\vspace{-2mm}
\mathop{\mathrm{min}}_{\bm{a},\bm{c}} \int_{-\infty}^{\infty} (h(x) - \widetilde{h}_{\bm{a},\bm{c}}(x))^2 \mathrm{d}x.
\end{equation}
We use the simulated annealing algorithm~\cite{kirkpatrick1983optimization} (more details in~\cref{appendix:regelu2}) to find a quasi-optimal weight $\bm{a}^*$ and bias $\bm{c}^*$. That is, for GELU we have
\vspace{-1.5mm}
\begin{equation*}
\begin{aligned}
\bm{a}^*_{gelu} &= [-0.04922, 1.098]^\top,\\
\bm{c}^*_{gelu} &= [-3.186, -0.001179, 3.191]^\top.
\end{aligned}
\vspace{-1.5mm}
\end{equation*}
And for SiLU we have
\vspace{-1.5mm}
\begin{equation*}
\begin{aligned}
\bm{a}^*_{silu} &= [-0.04060, 1.081]^\top,\\
\bm{c}^*_{silu} &= [-6.305, -0.0008685, 6.326]^\top.
\end{aligned}
\vspace{-1.5mm}
\end{equation*}
We denote the combination of GELU and $\mathrm{d}\widetilde{h}_{\bm{a}^*_{gelu}, \bm{c}^*_{gelu}}$ (or SiLU and $\mathrm{d}\widetilde{h}_{\bm{a}^*_{silu}, \bm{c}^*_{silu}}$) by ReGELU2 (or ReSiLU2).
Since ReGELU2 (or ReSiLU2) keeps the same primitive function as GELU (or SiLU), the initialization of the fine-tuning model is the exact pretrained model with GELU (or SiLU) activation function.
The main advantage of ReGELU2 and ReSiLU2 over GELU and SiLU, respectively, is that ReGELU2 and ReSiLU2 only need to store 2-bit activation for backward pass.
Our ReGELU2 and ReSiLU2 do not degrade the training efficiency, since they do not need extra computation for data range estimation~\cite{pan2021mesa}.
In addition, while setting a larger $k$ in \eqref{eq:relus_family} is also feasible for solving \eqref{eq:objective} using SGD, this will result in more memory and computational overhead. Since ReGELU2 and ReSiLU2 achieve comparable performance to GELU and SiLU in \cref{sec:experiments}, we recommend $k=2$ to be a universal choice.

\section{Memory-Sharing Backpropagation}
\label{sec:memory-shared_bp}
An insight on regular BP \eqref{eq:backward_pass} is that there exists redundancy when we store all $\{\bm{z}^i\}_{i=1}^L$ into activation memory.
To show this, we give a more detailed analysis on backward pass at the $i$-th layer $\bm{z}^{i} = \bm{h}^i(\bm{z}^{i-1}, \bm{\theta}^i)$.
In general, the purpose of backward pass at this layer is to calculate the gradient of the feature input $\frac{\partial \ell}{\partial \bm{z}^{i-1}}$ and the gradient of the parameter input $\frac{\partial \ell}{\partial \bm{\theta}^i}$ from the gradient of the feature output $\frac{\partial \ell}{\partial \bm{z}^i}$.
These calculations can be expressed in a general form as
\vspace{-1mm}
\begin{equation}
\begin{aligned}
\frac{\partial \ell}{\partial \bm{\theta}^i} &= \frac{\partial \bm{h}^i(\bm{z}^{i-1}, \bm{\theta}^i)}{\partial \bm{\theta}^i} \frac{\partial \ell}{\partial \bm{z}^i}, \\
\frac{\partial \ell}{\partial \bm{z}^{i-1}} &= \frac{\partial \bm{h}^i(\bm{z}^{i-1}, \bm{\theta}^i)}{\partial \bm{z}^{i-1}} \frac{\partial \ell}{\partial \bm{z}^i},
\end{aligned}
\vspace{-1mm}
\end{equation}
where $\frac{\partial \bm{h}^i(\bm{z}^{i-1}, \bm{\theta}^i)}{\partial \bm{\theta}^i}$ and $\frac{\partial \bm{h}^i(\bm{z}^{i-1}, \bm{\theta}^i)}{\partial \bm{z}^{i-1}}$ are the Jacobian matrices of $\bm{h}^i$ \textsl{w.r.t.} $\bm{\theta}^i$ and $\bm{z}^{i-1}$, respectively.
The reason for storing $\bm{z}^{i-1}$ into activation memory is that $\frac{\partial \bm{h}^i(\bm{z}^{i-1}, \bm{\theta}^i)}{\partial \bm{\theta}^i}$ and $\frac{\partial \bm{h}^i(\bm{z}^{i-1}, \bm{\theta}^i)}{\partial \bm{z}^{i-1}}$ involve the term $\bm{z}^{i-1}$.
However, this involvement is not always necessary.
In this section, we discuss about the situation in which the Jacobian matrices do not involve the term $\bm{z}^{i-1}$, and show how to use this property to achieve memory-sharing backpropagation (MS-BP) for avoiding the activation memory redundancy.

\subsection{Sufficient Condition of MS-BP}
\label{subsec:sufficient_condition_of_ms-bp}
We begin with a proposition about when the layer $\bm{h}^{i-1}$ can share the activation memory with the following layer $\bm{h}^i$.
\begin{proposition}
If the layer $\bm{h}^i$ satisfies the following conditions, we can reduce the activation memory in $\bm{h}^i$ by sharing its activation memory with $\bm{h}^{i+1}$:
\vspace{-1.5mm}
\begin{enumerate}
\vspace{-1.5mm}
\item $\bm{h}^i$ does not involve parameters $\bm{\theta}^i$, i.e.,
$\bm{z}^{i} = \bm{h}^i(\bm{z}^{i-1})$.
\vspace{-3mm}
\item The Jacobian matrix $\frac{\partial \bm{h}^i(\bm{z}^{i-1})}{\partial \bm{z}^{i-1}}$ can be reformulated as $\bm{J}(\bm{z}^{i}, \bm{\phi}^i)$, where $\bm{\phi}^i\in\mathbb{R}^{q_i}$ is an auxiliary variable with dimension $q_i \ll p_{i-1}$.
\vspace{-1.5mm}
\item The backward pass at $\bm{h}^{i+1}$ involves $\bm{z}^{i}$.
\end{enumerate}  
\vspace{-1.5mm}
\label{proposition:ms_bp}
\end{proposition}
Under the conditions in \cref{proposition:ms_bp}, the calculation of $\frac{\partial \ell}{\partial \bm{\theta}^i}$ is not required any more, and the calculation of $\frac{\partial \ell}{\partial \bm{z}^{i-1}}$ no longer needs $\bm{z}^{i-1}$.
Therefore, the intermediate feature $\bm{z}^{i-1}$ can be removed from the activation memory, and both $\bm{h}^i$ and $\bm{h}^{i+1}$ utilize $\bm{z}^{i}$ for gradient calculation.
Then the activation memory usage in $\bm{h}^{i}$ and $\bm{h}^{i+1}$ can be reduced from $\{\bm{z}^{i-1}, \bm{z}^{i}\}$ to $\{\bm{\phi}^i,\bm{z}^{i}\}$.
The first two conditions in \cref{proposition:ms_bp} are loose enough to cover simple element-wise activation functions and normalization layers.
But the third condition is not often met in fine-tuning networks when $\bm{h}^{i+1}$ is a frozen linear layer.
Unfortunately, the widely used SiLU does not satisfy the second condition (please refer to \cref{appendix:ms_act} for details). Thus, we mainly consider how to apply MS-BP to the layer normalization in \cref{sec:ms_norm}.

\subsection{Memory-Sharing Normalization}
\label{sec:ms_norm}
In this section, we describe the detailed technique for applying our MS-BP to LayerNorm~\cite{ba2016layer} and its variant RMSNorm~\cite{zhang2019root}.

\begin{algorithm}[t]
   \caption{Memory-Sharing Layer Normalization}
   \label{alg:memory-shared_norm}
\begin{algorithmic}
    \STATE Denote $\ell$ as the loss function.
    \STATE {\bfseries Input:} $\bm{z}^{i-1} \in \mathbb{R}^{p_{i-1}}$
    \STATE {\bfseries Forward Pass:}
    \STATE \quad $\sigma = \sqrt{p_{i-1}^{-1} {\bm{z}^{i-1}}^\top \bm{H}^\top \bm{H} \bm{z}^{i-1} + \varepsilon}$
    \STATE \quad $\bm{z}^i =  \sigma^{-1} \bm{H}\bm{z}^{i-1}$
    \STATE \quad \textsl{Save $\bm{z}^i$, $\sigma$ for backward pass}
    \STATE \quad {\bfseries Return Output:} $\bm{z}^i$
    \STATE {\bfseries Backward Pass:}
    \STATE \quad Receive gradient : $\frac{\partial \ell}{\partial \bm{z}^i}$
    \STATE \quad $\frac{\partial \ell}{\partial \bm{z}^{i-1}} = \sigma^{-1} \bm{H}^\top (\mathbbm{I} - p_{i-1}^{-1} \bm{z}^{i} {\bm{z}^{i}}^\top) \frac{\partial \ell}{\partial \bm{z}^i}$
    \STATE \quad {\bfseries Return Gradient:} $\frac{\partial \ell}{\partial \bm{z}^{i-1}}$
\end{algorithmic}
\end{algorithm}

The forward pass at the LayerNorm or RMSNorm and the following linear layer can be expressed as:
\vspace{-1.5mm}
\begin{equation}
\begin{aligned}
\sigma &=  \sqrt{p_{i-1}^{-1} {\bm{z}^{i-1}}^\top \bm{H}^\top \bm{H} \bm{z}^{i-1} + \varepsilon},
\\
\widetilde{\bm{z}}^{i-1} &=  \sigma^{-1} \bm{H}\bm{z}^{i-1},
\\
\bm{z}^i &= {\rm diag}(\bm{\alpha})\widetilde{\bm{z}}^{i-1} + \bm{\beta}, \\
\bm{z}^{i+1} &= \bm{W}\bm{z}^i + \bm{b},
\end{aligned}
\label{eq:norm_linear}
\vspace{-1.5mm}
\end{equation}
\vspace{-0mm}
where $\bm{H}$ is a general matrix. For LayerNorm we have $\bm{H} = \mathbbm{I} - p_{i-1}^{-1}\mathbbm{1}\mathbbm{1}^\top$, while for RMSNorm we have $\bm{H} = \mathbbm{I}, \ \bm{\beta} = \bm{0}$.
Here, $\mathbbm{I}$ is the identity matrix and $\mathbbm{1}$ is a vector of all ones.
$\varepsilon$ is a small positive scalar of $10^{-6}$ or $10^{-8}$.
$\bm{\alpha}$ and $\bm{\beta}$ are the affine weight and bias, respectively, in LayerNorm.

To satisfy the conditions in \cref{proposition:ms_bp}, we merge the affine parameters $\bm{\alpha}$ and $\bm{\beta}$ into the linear layer in \eqref{eq:norm_linear} as
\vspace{-1.5mm}
\begin{equation}
% \vspace{-1.1mm}
\begin{aligned}
\widetilde{\bm{W}} = \bm{W}{\rm diag}(\bm{\alpha}),\
\widetilde{\bm{b}} = \bm{W}\bm{\beta} + \bm{b}.
\end{aligned}
\vspace{-1.5mm}
\end{equation}
% 解释：becomes ----> is simplified as 4步变3步，可认为simplified
Then the forward pass is simplified as:
\vspace{-1.5mm}
\begin{equation}
\begin{aligned}
\sigma &= \sqrt{p_{i-1}^{-1} {\bm{z}^{i-1}}^\top \bm{H}^\top \bm{H} \bm{z}^{i-1} + \varepsilon},
\\
\bm{z}^{i} &= \sigma^{-1} \bm{H}\bm{z}^{i-1},
\\
\bm{z}^{i+1} &= \widetilde{\bm{W}}\bm{z}^{i} + \widetilde{\bm{b}}.
\end{aligned}
\vspace{-1.5mm}
\end{equation}
Now, we check the conditions in \cref{proposition:ms_bp}.
The first condition is met since there is no parameter in layer normalization after merging affine parameters.
The third condition is met at least in full tuning and LoRA, where the query and value projections are always adapted.\ To show the second condition is also met, we demonstrate how to reformulate the Jacobian matrix of the layer normalization in \cref{alg:memory-shared_norm}.
By this way, the total activation memory usage of a memory-sharing layer normalization and the following linear layer becomes the memory size of one vector in $\mathbb{R}^{p_{i-1}}$ and one scalar in $\mathbb{R}$.
We denote the memory-sharing LayerNorm as MS-LN and the memory-sharing RMSNorm as MS-RMSNorm (please refer to \cref{appendix:merge_rmsn} for details).

\section{Experiments}
\label{sec:experiments}

In this section, we conduct experiments by deploying our ReGELU2, ReSiLU2, MS-LN, and MS-RMSNorm into the representative ViT \cite{dosovitskiy2020vit} for vision tasks, as well as LLaMA \cite{touvron2023llama} and RoBERTa \cite{liu2019roberta} for natural language understanding tasks.
Specifically, we deploy our ReGELU2 (or ReSiLU2) into ViT, RoBERTa (or LLaMA) to replace the GELU (or SiLU) function.
MS-LN (or MS-RMSNorm) is also used to replace LayerNorm (or RMSNorm) with merged weights of pretrained ViT, RoBERTa (or LLaMA). Our method needs no extra operation in practical implementation.
% For training efficiency, we implement compatible CUDA kernels for our ReGELU2, ReSiLU2, MS-LN, and MS-RMSNorm and use FlashAttention \cite{dao2022flashattention} in all experiments.
We implement compatible CUDA kernels for our ReGELU2, ReSiLU2, MS-LN, and MS-RMSNorm.
FlashAttention \cite{dao2022flashattention} is used in the ViT and LLaMA experiments.
More experiments are put in \cref{appendix:more_results}.

\setlength{\tabcolsep}{4.0pt}
\begin{table*}[ht]
\vspace{-1mm}
\caption{\textbf{Average results on CIFAR10/100 and FGVC by fine-tuning ViT-base}. The best results are highlighted in \textbf{bold}.
%$1 \rm{MiB} = 1024^2 \rm{Bytes}$.
}
\centering
\begin{adjustbox}{max width =1.0\linewidth}
\begin{tabular}{cll|cll|cll}
\toprule
& & & \multicolumn{3}{c|}{Adapt Q, V} & \multicolumn{3}{c}{Adapt All Linear} \\
Method & Activation & Norm & Top-1(\%) & Mem.(MiB) & Thr.(images/s) & Top-1(\%) & Mem.(MiB) & Thr.(images/s) \\
\midrule
\multirow{7}{*}{\shortstack{LoRA\\$r = 4$}} & GELU & LN & \textbf{90.3} & 3827 & 288 & 90.7 & 5128 & \textbf{207}\\
 & Mesa-GELU & LN & \textbf{90.3} & 3453(-10\%) & 245(-15\%) & \textbf{90.8} & 4721(-8\%) & 186(-10\%) \\
 &\cellcolor{Gray}ReGELU2 &\cellcolor{Gray}LN &\cellcolor{Gray}\textbf{90.3} &\cellcolor{Gray}\textbf{3087}(-19\%) &\cellcolor{Gray}\textbf{289}(+0\%) &\cellcolor{Gray}\textbf{90.8} &\cellcolor{Gray}\textbf{4380}(-15\%) &\cellcolor{Gray}\textbf{207}(+0\%)\\
\cmidrule{2-9}
 & GELU & Mesa-LN & 90.2 & \textbf{3249}(-15\%) & 257(-11\%) & 90.8 & 4530(-12\%) & 189(-9\%)\\
 &\cellcolor{Gray}GELU &\cellcolor{Gray}MS-LN &\cellcolor{Gray}\textbf{90.7} &\cellcolor{Gray}3441(-10\%) &\cellcolor{Gray}\textbf{288}(+0\%) &\cellcolor{Gray}\textbf{91.2} &\cellcolor{Gray}\textbf{4316}(-16\%) &\cellcolor{Gray}\textbf{207}(+0\%)\\
\cmidrule{2-9}
 & Mesa-GELU & Mesa-LN & 90.4 & 2853(-25\%) & 226(-22\%) & 90.8 & 4209(-18\%) & 173(-17\%)\\
 &\cellcolor{Gray}ReGELU2 &\cellcolor{Gray}MS-LN &\cellcolor{Gray}\textbf{90.5} &\cellcolor{Gray}\textbf{2717}(-29\%) &\cellcolor{Gray}\textbf{290}(+1\%) &\cellcolor{Gray}\textbf{91.2} &\cellcolor{Gray}\textbf{3601}(-30\%) &\cellcolor{Gray}\textbf{208}(+0\%)\\
\midrule
\multirow{4}{*}{\shortstack{LoRA-FA\\$r = 4$}} & GELU & LN & \textbf{90.0} & 3386 & 304 & \textbf{90.2} & 3430 & 249\\
 & Mesa-GELU & LN & 89.9 & 3012(-11\%) & 261(-14\%) & \textbf{90.2} & 3021(-12\%) & 218(-12\%)\\
 & Mesa-GELU & Mesa-LN & 89.9 & \textbf{2411}(-29\%) & 236(-22\%) & 90.1 & \textbf{2457}(-28\%) & 200(-20\%)\\
 &\cellcolor{Gray}ReGELU2 &\cellcolor{Gray}LN &\cellcolor{Gray}89.8 &\cellcolor{Gray}2597(-23\%) &\cellcolor{Gray}\textbf{306}(+1\%) &\cellcolor{Gray}\textbf{90.2} &\cellcolor{Gray}2717(-21\%) &\cellcolor{Gray}\textbf{251}(+0\%)\\
\bottomrule
\end{tabular}
\end{adjustbox}
\label{tab:vit-b_lora}
% \vspace{-5mm}
\end{table*}

\setlength{\tabcolsep}{4.0pt}
\begin{table*}[ht]
\vspace{-1mm}
\caption{\textbf{Average results on CIFAR10/100 and FGVC by fine-tuning ViT-base and ViT-large}. The best results are highlighted in \textbf{bold}.
%$1 \rm{GiB} = 1024^3 \rm{Bytes}$.
}
\centering
\begin{adjustbox}{max width =1.0\linewidth}
\begin{tabular}{cll|cll|cll}
\toprule
& & & \multicolumn{3}{c|}{ViT-base} & \multicolumn{3}{c}{ViT-large} \\
Method & Activation & Norm & Top-1(\%) & Mem.(GiB) & Thr.(images/s) & Top-1(\%) & Mem.(GiB) & Thr.(images/s) \\
\midrule
\multirow{4}{*}{Full Tuning} & GELU & LN & 89.23 & 5.6 & 235 & 90.99 & 15.7 & 175 \\
 & ReGELU2 & LN & \textbf{89.31} & 4.9(-13\%) & 232(-1\%) & \textbf{91.15} & 13.7(-13\%) & 176(1\%) \\
 & GELU & MS-LN & 88.69 & 4.9(-14\%) & 238(+1\%) & 90.62 & 13.5(-14\%) & 182(4\%) \\
 & \cellcolor{Gray}ReGELU2 & \cellcolor{Gray}MS-LN & \cellcolor{Gray}88.75 & \cellcolor{Gray}\textbf{4.1}(-27\%) & \cellcolor{Gray}\textbf{241}(+2\%) & \cellcolor{Gray}90.96 & \cellcolor{Gray}\textbf{11.5}(-27\%) & \cellcolor{Gray}\textbf{183}(4\%) \\
\bottomrule
\end{tabular}
\end{adjustbox}
\label{tab:vit_full_tuning}
\vspace{-1mm}
\end{table*}

\setlength{\tabcolsep}{4.0pt}
\begin{table*}[h!]
\vspace{-1mm}
\caption{\textbf{Main results on fine-tuning LLaMA-7B and LLaMA-13B using QLoRA on Alpaca}.
``*'' indicates that the values are reported in QLoRA paper.
The best results are highlighted in \textbf{bold}.
$1 \rm{GiB} = 1024 \rm{MiB} = 1024^3 \rm{Bytes}$.
}
\centering
\begin{adjustbox}{max width =1.0\linewidth}
\begin{tabular}{cll|lll|lll}
\toprule
& & & \multicolumn{3}{c|}{LLaMA-7B} & \multicolumn{3}{c}{LLaMA-13B} \\
Method & Activation & Norm & Accuracy(\%) & Mem.(GiB) & Thr.(samples/s) & Accuracy(\%) & Mem.(GiB) & Thr.(samples/s) \\
\midrule
No Tuning & SiLU & RMSNorm & 35.65(35.1*) & & & 45.26(46.9*)\\
\midrule
\multirow{4}{*}{\shortstack{QLoRA\\$r = 64$\\All Linear}} & SiLU & RMSNorm & \textbf{40.75}(39.0*) & 20.6 & 7.9 & \textbf{46.68}(47.5*) & 31.4 & 5.8 \\
 & ReSiLU2 & RMSNorm & 39.86 & 19.0(-8\%) & 7.9(+0\%) & 46.59 & 29.0(-8\%) & 5.7(-2\%) \\
 & SiLU & MS-RMSNorm & 40.13 & 18.0(-12\%)  & 8.2(+3\%) & 46.34 & 27.5(-12\%) & 5.8(+0\%) \\
 & \cellcolor{Gray}ReSiLU2 & \cellcolor{Gray}MS-RMSNorm & \cellcolor{Gray}40.35 & \cellcolor{Gray}\textbf{14.6}(-29\%) & \cellcolor{Gray}\textbf{8.6}(+9\%) & \cellcolor{Gray}46.54  & \cellcolor{Gray}\textbf{22.3}(-29\%) & \cellcolor{Gray}\textbf{6.5}(+13\%) \\
\bottomrule
\end{tabular}
\end{adjustbox}
\label{tab:llama-7b-13b}
\vspace{-1mm}
\end{table*}

\setlength{\tabcolsep}{4.0pt}
\begin{table*}[h!]
\vspace{-1mm}
\caption{\textbf{Main results on fine-tuning RoBERTa-base using LoRA on GLUE}.
The best results are highlighted in \textbf{bold}.
}
\centering
\begin{adjustbox}{max width =1.0\linewidth}
\begin{tabular}{cll|lllll|cll}
\toprule
& & & \multicolumn{5}{c|}{Tasks} & \multicolumn{3}{c}{Mean} \\
Method & Activation & Norm & CoLA & SST-2 & MRPC & STS-B & RTE & Accuracy(\%) & Mem.(MiB) & Thr.(samples/s) \\
\midrule
\multirow{4}{*}{\shortstack{LoRA\\$r = 64$\\Q, K}}
& GELU    & LN    & 61.08          & 93.81          & 86.52          & 89.18          & 71.48          & 80.41          & 6517                 & \textbf{202}\\
& ReGELU2 & LN    & 58.03          & 93.46          & 87.75          & \textbf{89.73} & 69.31          & 79.66          & 5438(-17\%)          & 202(-0\%)\\
& GELU    & MS-LN & 57.52          & 94.04          & 86.52          & 89.18          & \textbf{75.45} & 80.54          & 6253(-4\%)           & 196(-3\%)\\
& \cellcolor{Gray}ReGELU2 & \cellcolor{Gray}MS-LN & \cellcolor{Gray}\textbf{61.60} & \cellcolor{Gray}\textbf{94.27} & \cellcolor{Gray}\textbf{87.99} & \cellcolor{Gray}89.71          & \cellcolor{Gray}75.09          & \cellcolor{Gray}\textbf{81.73} & \cellcolor{Gray}\textbf{5173}(-21\%) & \cellcolor{Gray}198(-2\%)\\
\bottomrule
\end{tabular}
\end{adjustbox}
\label{tab:roberta-base}
\vspace{-1mm}
\end{table*}

\subsection{Fine-Tuning ViT on Image Classification}
\label{exp:vit}
\textbf{Benchmark}.
Here, we employ the transformer models ViT-base and ViT-large pretrained on ImageNet-22k \cite{imagenet,dosovitskiy2020vit} as the backbones, which are fine-tuned on the CIFAR10/100 \cite{krizhevsky2009learning} and FGVC \cite{jia2022vpt} datasets. GELU and LayerNorm are the default modules in ViT-base and ViT-large.

\textbf{Fine-tuning}.
We implement our method with LoRA \cite{hu2022lora}, LoRA-FA \cite{zhang2023lorafa}, and full fine-tuning (Full-Tuning).
For LoRA, we adapt the weights of query and value projection or all linear layers.
Since the linear layers in LoRA-FA only store $\bm{A}\bm{x}$ instead of $\bm{x}$ in backward pass, our MS-LN can not reduce the activation memory usage to the following linear layers.
Therefore, we only use ReGELU2 for LoRA-FA in our experiments.
Please refer to \cref{appendix:implementation_details} for more implementation details.

\begin{figure}[t]
\begin{center}
\centerline{\includegraphics[width=\columnwidth]{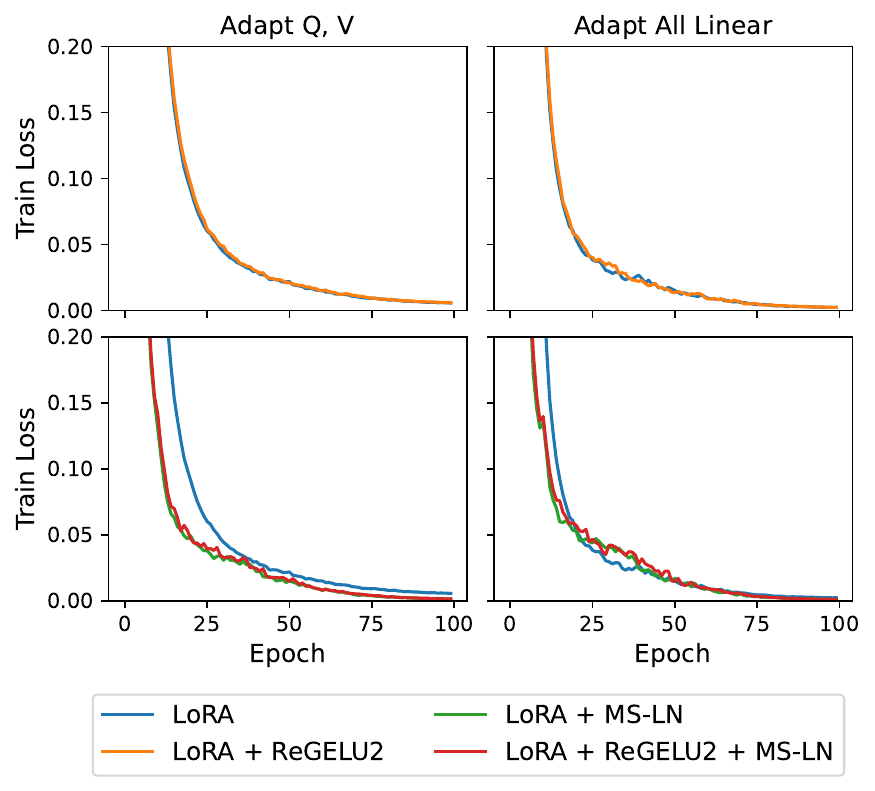}}
\vskip -0.2in
\caption{
\textbf{Convergence of ReGELU2 and MS-LN when using LoRA (rank $=4$) on ViT-base~\cite{dosovitskiy2020vit}}. The training loss is the average over the training loss on CIFAR10/100~\cite{krizhevsky2009learning} and FGVC~\cite{jia2022vpt}.
}
\label{fig:vit_base_convergence}
\end{center}
\vskip -0.4in
\end{figure}

\textbf{Comparison methods}.
% activation quantization
We compare our method with Mesa \cite{pan2021mesa}, a activation quantization method providing 8-bit GELU and LN.
We do not evaluate ActNN~\cite{chen2021actnn} since it is designed for CNNs and not usable to GELU and LN in ViTs. We also do not evaluate GACT~\cite{liu2022gact} due to training collapse in our experiments.

\textbf{Results}.\ In \cref{fig:vit_base_convergence}, we plot the average loss curves of fine-tuning ViT-base with LoRAs on CIFAR10/100~\cite{krizhevsky2009learning} and FGVC~\cite{jia2022vpt}.
We observe that the convergence tendency of our ReGELU2 is almost identical to that of GELU, while the training loss of ViT-base with our MS-LN decreases more rapidly than that without it.
This indicates that ReGELU2 preserves the learning capability of GELU while MS-LN accelerates the convergence speed.

In \cref{tab:vit-b_lora} and  \cref{tab:vit_full_tuning}, we compare the results of inference accuracy, activation memory usage, and training throughput, on fine-tuning ViT-base by LoRA, LoRA-FA, and Full-Tuning, respectively.
We observe that the LoRA with our method (ReGELU2 + MS-LN) reduces the peak GPU memory usage by $\sim$$1.1$ GiB and $\sim$$1.5$ GiB when adapting query/value projection and all linear layers, respectively, both occupying $\sim$$ 30 \%$ of peak GPU memory usage by vanilla LoRA.
Similarly, our method reduces $\sim$$27\%$ of the peak GPU memory usage in Full-Tuning.
In LoRA-FA fine-tuning, our ReGELU2 reduces the peak GPU memory usage by $\sim$$20\%$.
Besides, our method (ReGELU2 + MS-LN) does not degrade the training throughput and inference accuracy, while Mesa degrades training throughput clearly.

Since its activation memory behavior is independent of the fine-tuning methods, our ReGELU2 achieves consistent activation memory reduction in all cases of our experiments.
Due to frozen FFN modules in LoRA, the memory usage reduction by our MS-LN on adapting query and value projections is less than those on adapting all linear layers in LoRA and full tuning. Here, the third condition of \cref{proposition:ms_bp} is not satisfied for the LN in the FFN modules.
%

% The convergence speed of ViT-base with our MS-LN is usually faster than that with vanilla LN.
% This indicates a stronger learning capability of ViT-base with our MS-LN, yet with higher propensity of overfitting, than that without our MS-LN.
% For instance, the Top-1 accuracies of fine-tuning ViT-base by LoRA with our MS-LN consistently surpass those with the vanilla LN, while the tendency is reversed on fine-tuning ViT-base by Full-Tuning (\cref{tab:vit_full_tuning}).
% 下一句对实验描述没帮助，还会让审稿人问我们为什么不做相关实验
% Fortunately, the side effect of overfitting can always be mitigated by increasing the weight decay and reducing the learning rate.

% Natural
\subsection{Fine-Tuning LLaMA on Language Understanding}
\label{exp:llama}

\textbf{Benchmark}. We fine-tune LLaMA-7B and LLaMA-13B
\\ % please do not remove it
\cite{touvron2023llama} using Alpaca \cite{taori2023stanford} and evaluate the fine-tuned models on 5-shot MMLU \cite{hendrycks2020measuring}.
LLaMA uses SwiGLU~\cite{shazeer2020glu} (containing SiLU in its implementation) for activation and RMSNorm for layer normalization.
The training uses model parallel provided in the Transformers package \cite{wolf2020transformers} with 2$\times$H800 GPUs. The reported peak memory usage is the max value of those from the 2 GPUs.

\textbf{Fine-tuning}. We deploy our method into QLoRA~\cite{dettmers2023qlora} to fine-tune LLaMA-7B and LLaMA-13B.
QLoRA uses NF4 data type to store the pretrained weights and uses Bfloat16 to store the parameters in LoRA.
In QLoRA, all projection weights in linear layers are adapted by LoRA.
When applying our MS-RMSNorm to merge the affine parameters, we transpose the weight matrix of the pretrained parameters, to avoid changing the conditional distribution of the block-wise quantization in QLoRA.
Please refer to \cref{appendix:implementation_details} for more implementation details.
% %
% \textbf{Implementation details:}
% %
% The batch size is $4$ and the number of gradient accumulation steps is $4$.
% %
% We tune the constant learning rate in \{1e-4, 2e-4\} and report the best 5-shot MMLU accuracy among them.
% %
% The training uses Model Parallel (MP) provided by \rm{Huggingface} with two NVIDIA H800.

\textbf{Results} on fine-tuning LLaMA-7B and LLaMA-13B are summarized in \cref{tab:llama-7b-13b}.
We observe that fine-tuning LLaMAs by our method achieves comparable MMLU accuracy to the baseline.
Our method substantially reduces the peak memory usage on fine-tuning LLaMAs by QLoRA, \ie, $\sim$$ 6.0$ GiB on fine-tuning LLaMA-7B and $\sim$$ 9.1$ GiB on fine-tuning LLaMA-13B, representing a significant amount of GPU memory savings.
The reduction amounts both occupy $\sim$$ 30\%$ of the baseline's peak GPU memory usage.
What’s more, our method yields an $\sim$$ 10\%$ improvement of training throughput on fine-tuning LLaMA-7B and LLaMA-13B with QLoRA.
Fine-tuning LLaMA-7B and LLaMA-13B with our method suffer from slight accuracy drops of $0.40\%$ and $0.14\%$, respectively. This indicates that our method can be potentially applied to larger transformers.

Note that fine-tuning LLaMAs with both ReSiLU2 and MS-RMSNorm achieves larger memory usage reduction than the sum of reductions by using them separately. This is possibly attributed to the implementation details of QLoRA.
% the quantilization process

\subsection{Fine-Tuning RoBERTa on Language Understanding}
\label{exp:roberta-base}

\textbf{Benchmark}. We fine-tune the pretrained RoBERTa-base \cite{liu2019roberta} on five taskes of GLUE \cite{wang2018glue}, \ie, CoLA, SST-2, MRPC, STS-B and RTE.
RoBERTa-base uses GELU and LayerNorm.
The training uses model parallel provided in the Transformers package~\cite{wolf2020transformers} with 2$\times$RTX4090 GPUs. The reported usage
of peak memory overhead is the sum of those  from the 2 GPUs.

\textbf{Fine-tuning}. We implement our method with LoRA to fine-tune the pretrained RoBERTa-base.
The data type in this experiment is FP32.
Please refer to \cref{appendix:implementation_details} for more implementation details.

\textbf{Results} on fine-tuning RoBERTa-base are summarized in \cref{tab:roberta-base}.
Fine-tuning RoBERTa-base with our method achieves comparable accuracy and training throughput to the baseline.
Our method reduces the amount of GPU memory usage by $\sim$$21\%$.
Here, MS-LN gets less reduction of GPU memory usage than ReGELU2, which may be attributed to two reasons. First, we use FP32 in this experiment, so that LayerNorm occupies less proportion of activation memory usage than that in AMP training. Secondly, since LoRA only adapts projection weights in the queries and keys in the attention modules, the third condition of \cref{proposition:ms_bp} is not satisfied for the LN in the FFN modules.

\section{Conclusion}
To reduce the activation memory overhead in backpropagation (BP), in this paper, we introduced an Approximate Backpropagation (Approx-BP) theory and a Memory-sharing Backpropagation (MS-BP) strategy.
Our Approx-BP theory revealed the feasibility of decoupling the primitive and derivative functions of network layers for training. We derived the ReGELU2 and ReSiLU2 as alternatives of the GELU and SiLU, respectively, used in transformers.
We applied our MS-BP strategy into layer normalization (LN), and proposed MS-LN (or MS-RMSNorm) to remove the activation memory redundancy between LN and the following linear layers in regular BP.
Experimental results demonstrated that our method reduces up to $\sim$$30\%$ of the peak GPU memory usage on fine-tuning transformers, with comparable accuracy and no drop on training throughput.
We believe that our method can be applied to not only fine-tuning stage but also pretraining stage.
Even though pretraining exceeds our research scope, we have explored how our method can benefit the pretraining from two aspects.
In \cref{appendix:llama_experiments}, we show that our method can increase the length of training sequence substantially.
In \cref{appendix:llama_experiments}, our method can reduce the communication times in the distributed training by allowing a large batch size, thereby increasing the training throughput significantly.
%
% Even though pretraining exceeds our research scope, we provide some results to support that our method can increase the length of training sequence and the training throughput (\cref{appendix:llama_experiments,appendix:bert_experiment}).

% \clearpage
\section*{Acknowledgements}
This work is supported in part by National Natural Science Foundation of China (No. 12226007 and 62176068), the Fundamental Research Funds for the Central Universities, and CAAI-Huawei MindSpore Open Fund.

\section*{Impact Statement}
This paper presents a work whose goal is to advance the field of machine learning. There are many potential societal consequences of our work, none of which we feel must be specifically highlighted here.
However, our work has the potential contribution to positively lowering the fine-tuning barrier of large models and promoting their popularity in both research community and industrial applications.

\bibliography{main}
\bibliographystyle{icml2024}

%%%%%%%%%%%%%%%%%%%%%%%%%%%%%%%%%%%%%%%%%%%%%%%%%%%%%%%%%%%%%%%%%%%%%%%%%%%%%%%
%%%%%%%%%%%%%%%%%%%%%%%%%%%%%%%%%%%%%%%%%%%%%%%%%%%%%%%%%%%%%%%%%%%%%%%%%%%%%%%
% APPENDIX
%%%%%%%%%%%%%%%%%%%%%%%%%%%%%%%%%%%%%%%%%%%%%%%%%%%%%%%%%%%%%%%%%%%%%%%%%%%%%%%
%%%%%%%%%%%%%%%%%%%%%%%%%%%%%%%%%%%%%%%%%%%%%%%%%%%%%%%%%%%%%%%%%%%%%%%%%%%%%%%
\newpage
\appendix
\onecolumn

\section{Qualitative Comparison of Related Works}
\label{appendix:qualitative_comparison_of_related_works}
In \cref{tab:mehtods_featurs}, we provide qualitative comparison of different methods on three aspects, \ie, applicable to non-linear layers (``Non-Linear''), keep training throughput (``Keep Throughput''), and applicable beyond LoRAs (``Beyond LoRA'').
Our method can reduce the activation memory usage in non-linear layers, which can not be achieved by parameter freezing techniques~\cite{hu2022lora,jia2022vpt} or LoRA-FA~\cite{zhang2023lorafa}.
One key advantage of our method over gradient checkpointing~\cite{chen2016training} and ACT methods~\cite{pan2021mesa,liu2022gact} is that our method does not degrade the training efficiency.
\begin{table}[h]
\vspace{-2mm}
  \begin{center}
    \caption{\textbf{Comparison of different methods on activation memory reduction.}
    ``Freeze'': freezing some parameters in fine-tuning.
    ``CKPT'': Gradient Checkpointing~\cite{chen2016training}.
    ``ACT'': Activation Compression Training~\cite{pan2021mesa,liu2022gact}.
    }
    % \vspace{-1.1mm}
    \label{tab:mehtods_featurs}
\begin{adjustbox}{max width =1.0\linewidth}
    \begin{tabular}{r|c|c|c}
      \toprule % <-- Toprule here
      Method & \shortstack{Non-Linear} & \shortstack{Keep Throughput} & \shortstack{Beyond LoRA} \\
      \midrule % <-- Midrule here
      Freeze & \xmark & \cmark & \cmark \\
      \midrule % <-- Midrule here
      CKPT~\cite{chen2016training} & \cmark & \xmark & \cmark \\
      \midrule % <-- Midrule here
      ACT~\cite{pan2021mesa,liu2022gact} & \cmark & \xmark & \cmark \\
      \midrule % <-- Midrule here
      LoRA-FA~\cite{zhang2023lorafa} & \xmark & \cmark & \xmark \\
      \midrule
      \textbf{Our Method} & \cmark & \cmark & \cmark \\
      \bottomrule % <-- Bottomrule here
    \end{tabular}
\end{adjustbox}
  \end{center}
  \vspace{-6mm}
\end{table}

\section{Analyses on activation memory allocation in each block of ViT and LLaMA}
\label{appendix:analysis_on_block}
\begin{figure*}[h!]
% \vskip -0.2in
\begin{center}
\includegraphics[width=\linewidth]{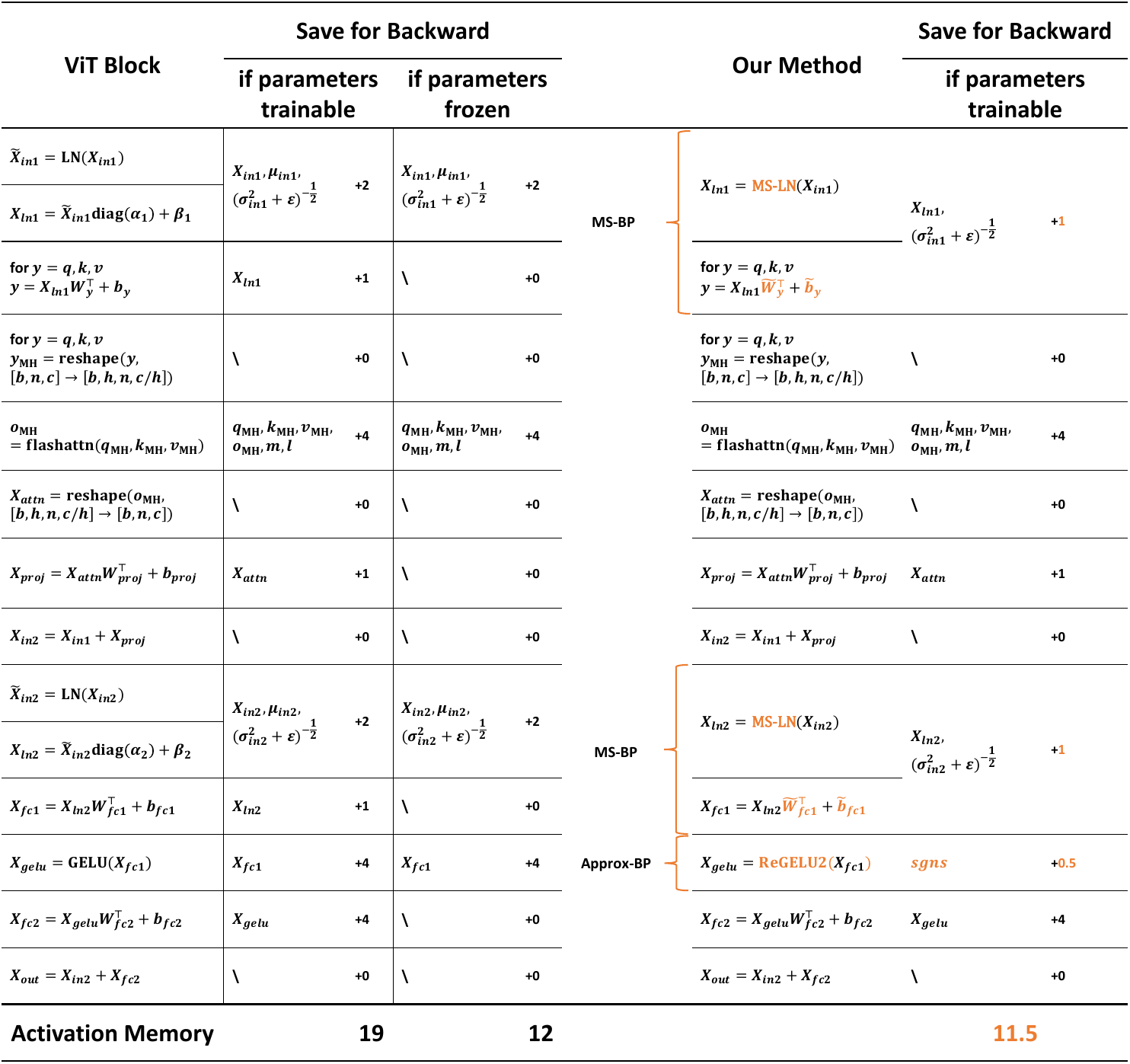}
%\vskip -0.2in
\caption{\textbf{Composition of the activation memory in each block of ViT~\cite{dosovitskiy2020vit}}. We assume Layer Normalization uses fp32, other operators use fp16 data type and each operator in the table is implemented as a single CUDA kernel. \textbf{The unit of memory is the memory size of a tensor (16 bits type) with the shape $[b,n,c]$}.}
\label{fig:vit_activation_memory}
\end{center}
%\vskip -0.2in
\end{figure*}

\begin{figure*}[h!]
% \vskip -0.2in
\begin{center}
\includegraphics[width=\linewidth]{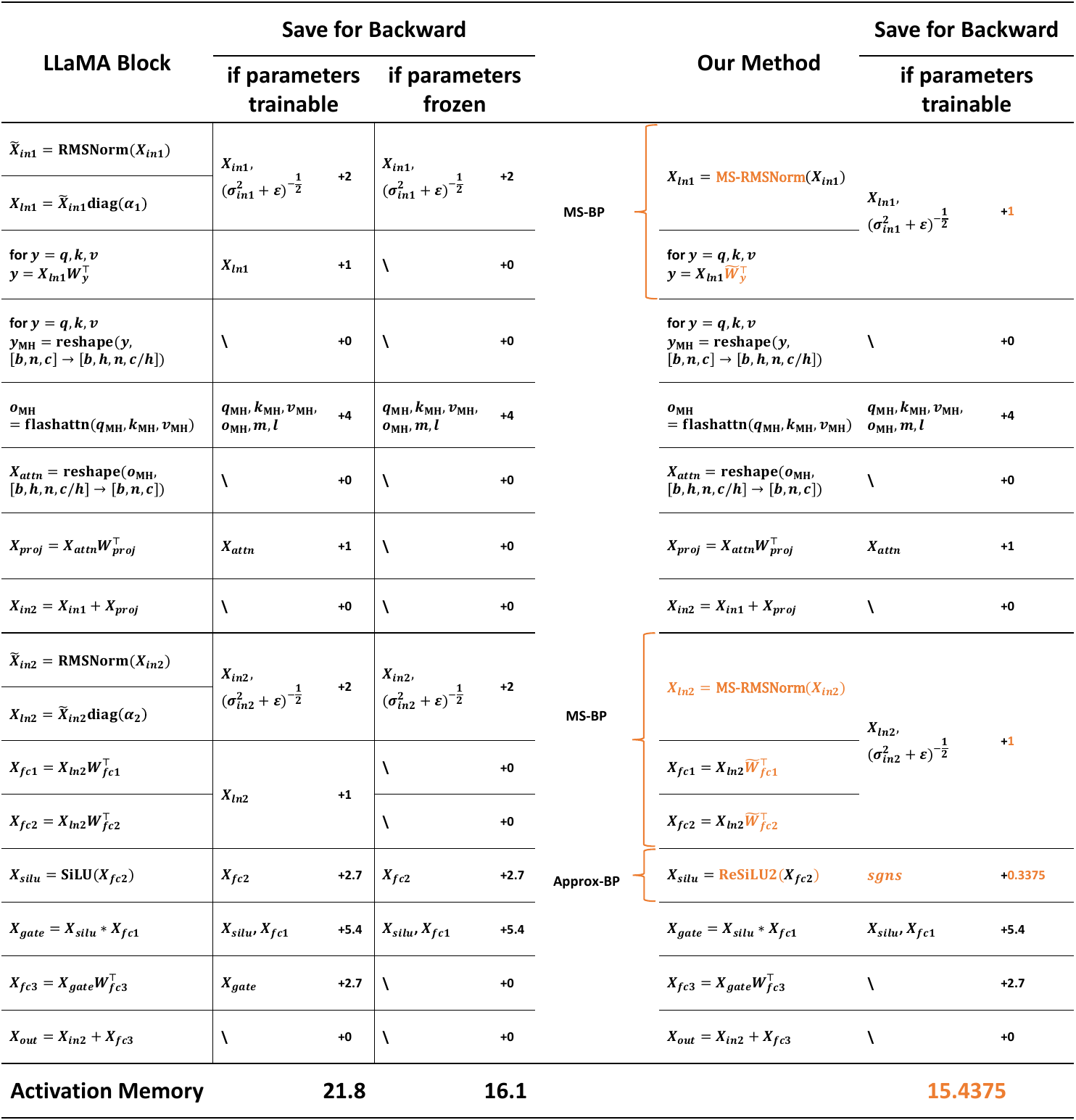}
%\vskip -0.2in
\caption{\textbf{Composition of activation memory in each block of LLaMA~\cite{touvron2023llama}}. Here, RMSNorm uses fp32, other operators use bf16 data type, and each operator is implemented as a single CUDA kernel. In practice, RMSNorm is often implemented by multiple sub-operators, which may bring additional memory usage. The unit of memory in this figure is the memory size of a tensor (16 bits type) with the shape $[b,n,c]$. The expanding factor in LLaMA depends on the model size, we use LLaMA-13B as an example.}
\label{fig:llama_activation_memory}
\end{center}
%\vskip -0.2in
\end{figure*}

We present detailed analysis of the activation memory allocation for each operator within the transformer blocks of ViT~\cite{dosovitskiy2020vit} and LLaMA~\cite{touvron2023llama}.
For ViT, refer to \cref{fig:vit_activation_memory}; for LLaMA, refer to \cref{fig:llama_activation_memory}.

\section{Possibility of Substituting the Forward Pass of Activation Funcition}
\label{appendix:possibility_of_sf}
We also investigate the possibility of changing the whole activation function including forward pass.
Nevertheless, empirical results show that changing forward pass of activation function severely degrades the fine-tuning performance.
We attribute this phenomenon to the criticality of model initialization.
Specifically, replacing SiLU by $\widetilde{h}_{\bm{a}^*_{silu}, \bm{c}^*_{silu}}(x)$ in \eqref{eq:relus_family}, the no-tuning MMLU accuracy of LLaMA-7B degrades from $35.62\%$ to $23.44\%$ and the no-tuning MMLU accuracy of LLaMA-13B degrades from $45.26\%$ to $23.51\%$.
Hence, we retain the forward pass in activation function.

\section{Proof of theorems}
\label{appendix:proof}
\begin{proof}[Proof of \cref{thm:abp}]
According to the definitions in \cref{subsec:approx-bp_theory}, we have the following decomposition:
\begin{equation}
\begin{aligned}
\Vert \widehat{\bm{g}} - \bm{g} \Vert_2
&= \Vert \widehat{\bm{g}} - \widetilde{\bm{g}} + \widetilde{\bm{g}} - \bm{g} \Vert_2
\leq \Vert \widehat{\bm{g}} - \widetilde{\bm{g}} \Vert_2 + \Vert \widetilde{\bm{g}} - \bm{g} \Vert_2 \\
&= \Vert \widetilde{\bm{g}}(\ell(\bm{z}^L), \bm{z}, \bm{\theta}) - \widetilde{\bm{g}}(\ell(\widetilde{\bm{z}}^L), \widetilde{\bm{z}}, \bm{\theta}) \Vert_2 + \Vert \frac{\partial}{\partial \bm{\theta}} \ell(\widetilde{\bm{f}}(\bm{x}, \bm{\theta})) - \frac{\partial}{\partial \bm{\theta}} \ell(\bm{f}(\bm{x}, \bm{\theta}))\Vert_2.
\end{aligned}
\label{eq:decomposition}
\end{equation}
By A1, $\exists a_1 > 0$, $\forall \bm{\theta}$, we have
\begin{equation}
\begin{aligned}
&\Vert \widetilde{\bm{g}}(\ell(\bm{z}^L), \bm{z}, \bm{\theta}) - \widetilde{\bm{g}}(\ell(\widetilde{\bm{z}}^L), \widetilde{\bm{z}}, \bm{\theta}) \Vert_2
\leq a_1 (\Vert \ell(\bm{z}^L) - \ell(\widetilde{\bm{z}}^L) \Vert_2 + \Vert \bm{z} - \widetilde{\bm{z}} \Vert_2).
\end{aligned}
\end{equation}
By A2, $\exists\ a_2 > 0$, such that
\begin{equation}
\begin{aligned}
&\Vert \ell(\bm{z}^L) - \ell(\widetilde{\bm{z}}^L) \Vert_2 \leq a_2\Vert \bm{z}^L - \widetilde{\bm{z}}^L \Vert_2.
\end{aligned}
\end{equation}
Combining the above inequalities, we have
\begin{equation}
\begin{aligned}
\Vert \widetilde{\bm{g}}(\ell(\bm{z}^L), \bm{z}, \bm{\theta}) - \widetilde{\bm{g}}(\ell(\widetilde{\bm{z}}^L), \widetilde{\bm{z}}, \bm{\theta}) \Vert_2
\leq& a_1a_2\Vert \bm{z}^L - \widetilde{\bm{z}}^L \Vert_2 + a_1\Vert \bm{z} - \widetilde{\bm{z}} \Vert_2 \\
\leq& (1 + a_1 a_2) \Vert \bm{z}^L - \widetilde{\bm{z}}^L \Vert_2 + a_1 \sum_{i=1}^{L-1} \Vert \bm{z}^{i} - \widetilde{\bm{z}}^{i} \Vert_2.
\end{aligned}
\label{eq:term_1}
\end{equation}
By A3, $\exists M_1 > 0$, $\exists M_2 > 0$, $\forall \bm{x}$, $\forall \bm{\theta} \in \mathbb{R}^M$, $\forall \bm{q} \in \mathbb{R}^M$, we have
\begin{equation}
\Vert \frac{\partial^2}{\partial \bm{\theta} \partial \bm{\theta}} \ell (\bm{f} (\bm{x}, \bm{\theta}) ) \bm{q} \Vert_2 \leq M_1 \Vert \bm{q} \Vert_2 
\quad \mathrm{and} \quad
\Vert \frac{\partial^2}{\partial \bm{\theta} \partial \bm{\theta}} \ell (\widetilde{\bm{f}} (\bm{x}, \bm{\theta}) ) \bm{q} \Vert_2 \leq M_1 \Vert \bm{q} \Vert_2.
\end{equation}

By Taylor expansion with Lagrange remainder, $\forall t \in (0, \infty)$ and $\forall \bm{q} \in \mathbb{R}^M$, we have
\begin{subequations}
\begin{align}
\ell(\widetilde{\bm{f}}(\bm{x}, \bm{\theta} + t \bm{q})) - \ell(\bm{f}(\bm{x}, \bm{\theta} + t \bm{q}))
= &\ell(\widetilde{\bm{f}}(\bm{x}, \bm{\theta})) - \ell(\bm{f}(\bm{x}, \bm{\theta})) + t \bm{q}^\top (\frac{\partial}{\partial \bm{\theta}} \ell(\widetilde{\bm{f}}(\bm{x}, \bm{\theta})) - \frac{\partial}{\partial \bm{\theta}} \ell(\bm{f}(\bm{x}, \bm{\theta}))) \notag \\
+ &\frac{t^2}{2} \bm{q}^\top (\frac{\partial^2}{\partial \bm{\theta} \partial \bm{\theta}} \ell(\widetilde{\bm{f}}(\bm{x}, \bm{\theta} + \xi_1 \bm{q})) - \frac{\partial^2}{\partial \bm{\theta} \partial \bm{\theta}} \ell(\bm{f}(\bm{x}, \bm{\theta} + \xi_1 \bm{q})))) \bm{q}, \label{eq:Taylor_1}\\
\ell(\widetilde{\bm{f}}(\bm{x}, \bm{\theta} - t \bm{q})) - \ell(\bm{f}(\bm{x}, \bm{\theta} - t \bm{q}))
= &\ell(\widetilde{\bm{f}}(\bm{x}, \bm{\theta})) - \ell(\bm{f}(\bm{x}, \bm{\theta})) - t \bm{q}^\top (\frac{\partial}{\partial \bm{\theta}} \ell(\widetilde{\bm{f}}(\bm{x}, \bm{\theta})) - \frac{\partial}{\partial \bm{\theta}} \ell(\bm{f}(\bm{x}, \bm{\theta}))) \notag \\
+ &\frac{t^2}{2} \bm{q}^\top (\frac{\partial^2}{\partial \bm{\theta} \partial \bm{\theta}} \ell(\widetilde{\bm{f}}(\bm{x}, \bm{\theta} - \xi_2 \bm{q})) - \frac{\partial^2}{\partial \bm{\theta} \partial \bm{\theta}} \ell(\bm{f}(\bm{x}, \bm{\theta} - \xi_2 \bm{q})))) \bm{q}, \label{eq:Taylor_2}
\end{align}
\end{subequations}
where $\xi_1, \xi_2 \in (0, t)$. $\frac{\partial^2}{\partial \bm{\theta} \partial \bm{\theta}} \ell(\widetilde{\bm{f}}(\bm{x}, \bm{\theta}))$ and $\frac{\partial^2}{\partial \bm{\theta} \partial \bm{\theta}} \ell(\bm{f}(\bm{x}, \bm{\theta}))$ are the Hessian matrices of $\widetilde{\bm{f}}(\bm{x}, \bm{\theta})$ and $\bm{f}(\bm{x}, \bm{\theta})$, respectively.
From \eqref{eq:Taylor_1} and \eqref{eq:Taylor_2}, we derive
\begin{equation}
\begin{aligned}
&\bm{q}^\top (\frac{\partial}{\partial \bm{\theta}} \ell(\widetilde{\bm{f}}(\bm{x}, \bm{\theta})) - \frac{\partial}{\partial \bm{\theta}} \ell(\bm{f}(\bm{x}, \bm{\theta}))) \\
=& \frac{1}{2t}(\ell(\widetilde{\bm{f}}(\bm{x}, \bm{\theta} + t\bm{q})) - \ell(\bm{f}(\bm{x}, \bm{\theta} + t\bm{q})) - \ell(\widetilde{\bm{f}}(\bm{x}, \bm{\theta} - t\bm{q})) + \ell(\bm{f}(\bm{x}, \bm{\theta} - t\bm{q}))) \\
+& \frac{t}{4} \bm{q}^\top (\frac{\partial^2}{\partial \bm{\theta} \partial \bm{\theta}} \ell(\widetilde{\bm{f}}(\bm{x}, \bm{\theta} - \xi_2 \bm{q})) - \frac{\partial^2}{\partial \bm{\theta} \partial \bm{\theta}} \ell(\bm{f}(\bm{x}, \bm{\theta} - \xi_2 \bm{q})) - \frac{\partial^2}{\partial \bm{\theta} \partial \bm{\theta}} \ell(\widetilde{\bm{f}}(\bm{x}, \bm{\theta} + \xi_1 \bm{q})) + \frac{\partial^2}{\partial \bm{\theta} \partial \bm{\theta}} \ell(\bm{f}(\bm{x}, \bm{\theta} + \xi_1 \bm{q}))) \bm{q} \\
\leq& \frac{1}{t} \sup_{\bm{\theta}} |\ell (\widetilde{\bm{f}}(\bm{x}, \bm{\theta})) - \ell (\bm{f}(\bm{x}, \bm{\theta}))| + \frac{t}{2}(M_1 + M_2) \bm{q}^\top \bm{q}.
\end{aligned}
\label{eq:Taylor_3}
\end{equation}
Since \eqref{eq:Taylor_3} is valid for all $\bm{q} \in \mathbb{R}^M$ and $t \in (0, \infty)$, by setting
\begin{equation}
\begin{aligned}
\bm{q} =& \frac{\partial}{\partial \bm{\theta}} \ell(\widetilde{\bm{f}}(\bm{x}, \bm{\theta})) - \frac{\partial}{\partial \bm{\theta}} \ell(\bm{f}(\bm{x}, \bm{\theta})), \\
t =& \sqrt{\frac{2 \sup_{\bm{\theta}} |\ell (\widetilde{\bm{f}}(\bm{x}, \bm{\theta})) - \ell (\bm{f}(\bm{x}, \bm{\theta}))|}{(M_1 + M_2) \Vert \frac{\partial}{\partial \bm{\theta}} \ell(\widetilde{\bm{f}}(\bm{x}, \bm{\theta})) - \frac{\partial}{\partial \bm{\theta}} \ell(\bm{f}(\bm{x}, \bm{\theta})) \Vert_2^2}},
\end{aligned}
\end{equation}
we have
\begin{equation}
\begin{aligned}
\Vert \frac{\partial}{\partial \bm{\theta}} \ell(\widetilde{\bm{f}}(\bm{x}, \bm{\theta})) - \frac{\partial}{\partial \bm{\theta}} \ell(\bm{f}(\bm{x}, \bm{\theta})) \Vert_2 
&\leq \sqrt{2(M_1 + M_2)} \sqrt{\sup_{\bm{\theta}}|\ell (\widetilde{\bm{f}}(\bm{x}, \bm{\theta})) - \ell (\bm{f}(\bm{x}, \bm{\theta}))|} \\
&= \sqrt{2(M_1 + M_2)a_2} \sqrt{\sup_{\bm{\theta}} \Vert \widetilde{\bm{z}}^L - \bm{z}^L \Vert_2}.
\end{aligned}
\label{eq:term_2}
\end{equation}
By A2, for $i=2,...,L$, $\exists b_i > 0$, $\forall \bm{\theta}^i$, we have
\begin{equation}
\Vert \bm{h}^i(\bm{z}^{i-1}, \bm{\theta}^i) - \bm{h}^i(\widetilde{\bm{z}}^{i-1}, \bm{\theta}^i) \Vert_2
\leq b_i \Vert \bm{z}^{i-1} - \widetilde{\bm{z}}^{i-1} \Vert_2.
\end{equation}
Therefore, we attain
\begin{equation}
\begin{aligned}
&\Vert \widetilde{\bm{z}}^i - \bm{z}^i \Vert_2 \\
= &\Vert \widetilde{\bm{h}}^{i}_{\bm{\theta}} \circ \widetilde{\bm{h}}^{i-1}_{\bm{\theta}} \circ ... \circ \widetilde{\bm{h}}^{1}_{\bm{\theta}} (\bm{x}) - \bm{h}^{i}_{\bm{\theta}} \circ \bm{h}^{i-1}_{\bm{\theta}} \circ ... \circ \bm{h}^{1}_{\bm{\theta}} (\bm{x}) \Vert_2 \\
\leq &\Vert \widetilde{\bm{h}}^{i}_{\bm{\theta}} \circ \widetilde{\bm{h}}^{i-1}_{\bm{\theta}} \circ ... \circ \widetilde{\bm{h}}^{1}_{\bm{\theta}} (\bm{x}) - \bm{h}^{i}_{\bm{\theta}} \circ \widetilde{\bm{h}}^{i-1}_{\bm{\theta}} \circ ... \circ \widetilde{\bm{h}}^{1}_{\bm{\theta}} (\bm{x}) \Vert_2 + \Vert \bm{h}^{i}_{\bm{\theta}} \circ \widetilde{\bm{h}}^{i-1}_{\bm{\theta}} \circ ... \circ \widetilde{\bm{h}}^{1}_{\bm{\theta}} (\bm{x}) - \bm{h}^{i}_{\bm{\theta}} \circ \bm{h}^{i-1}_{\bm{\theta}} \circ ... \circ \bm{h}^{1}_{\bm{\theta}} (\bm{x}) \Vert_2 \\
\leq &\sup_{\bm{z}^{i-1}} \Vert \widetilde{\bm{h}}^{i}(\bm{z}^{i-1}, \bm{\theta}^{i}) - \bm{h}^{i}(\bm{z}^{i-1}, \bm{\theta}^{i}) \Vert_2 + b_i \Vert \widetilde{\bm{h}}^{i-1}_{\bm{\theta}} \circ \widetilde{\bm{h}}^{i-2}_{\bm{\theta}} \circ ... \circ \widetilde{\bm{h}}^{1}_{\bm{\theta}} (\bm{x}) - \bm{h}^{i-1}_{\bm{\theta}} \circ \bm{h}^{i-2}_{\bm{\theta}} \circ ... \circ \bm{h}^{1}_{\bm{\theta}} (\bm{x}) \Vert_2 \\
\leq &\sup_{\bm{z}^{i-1}} \Vert \widetilde{\bm{h}}^{i}(\bm{z}^{i-1}, \bm{\theta}^{i}) - \bm{h}^{i}(\bm{z}^{i-1}, \bm{\theta}^{i}) \Vert_2 + b_i \sup_{\bm{z}^{i-2}} \Vert \widetilde{\bm{h}}^{i-1}(\bm{z}^{i-2}, \bm{\theta}^{i-1}) - \bm{h}^{i-1}(\bm{z}^{i-2}, \bm{\theta}^{i-1}) \Vert_2 \\
+ &... + b_i b_{i-1} ... b_2 \sup_{\bm{z}^0} \Vert \widetilde{\bm{h}}^{1}(\bm{z}^{0}, \bm{\theta}^{1}) - \bm{h}^{1}(\bm{z}^{0}, \bm{\theta}^{1}) \Vert_2.
\end{aligned}
\label{eq:term_3}
\end{equation}
From \eqref{eq:term_1} and \eqref{eq:term_3}, we derive that $\exists \alpha_1 > 0$, such that
\begin{equation}
\begin{aligned}
\Vert \widetilde{\bm{g}}(\ell(\bm{z}^L), \bm{z}, \bm{\theta}) - \widetilde{\bm{g}}(\ell(\widetilde{\bm{z}}^L), \widetilde{\bm{z}}, \bm{\theta}) \Vert_2
&\leq \alpha_1 \sum_{i=1}^L \sup_{\bm{z}^{i-1}} \Vert \widetilde{\bm{h}}^{i}(\bm{z}^{i-1}, \bm{\theta}^{i}) - \bm{h}^{i}(\bm{z}^{i-1}, \bm{\theta}^{i}) \Vert_2 \\
&\leq \alpha_1 \sum_{i=1}^L \sup_{\bm{z}^{i-1}, \bm{\theta}^{i}} \Vert \widetilde{\bm{h}}^{i}(\bm{z}^{i-1}, \bm{\theta}^{i}) - \bm{h}^{i}(\bm{z}^{i-1}, \bm{\theta}^{i}) \Vert_2.
\end{aligned}
\end{equation}
From \eqref{eq:term_2} and \eqref{eq:term_3}, we derive that $\exists \alpha_2 > 0$, such that
\begin{equation}
\begin{aligned}
\Vert \frac{\partial}{\partial \bm{\theta}} \ell(\widetilde{\bm{f}}(\bm{x}, \bm{\theta})) - \frac{\partial}{\partial \bm{\theta}} \ell(\bm{f}(\bm{x}, \bm{\theta})) \Vert_2 
&\leq \alpha_2 \sqrt{\sum_{i=1}^L \sup_{\bm{z}^{i-1}, \bm{\theta}^{i}} \Vert \widetilde{\bm{h}}^{i}(\bm{z}^{i-1}, \bm{\theta}^{i}) - \bm{h}^{i}(\bm{z}^{i-1}, \bm{\theta}^{i}) \Vert_2}.
\end{aligned}
\end{equation}
By \eqref{eq:decomposition} and setting $\alpha = \max \{\alpha_1, \alpha_2\}$, we attain
\begin{equation}
\Vert \widehat{\bm{g}} - \bm{g} \Vert_2
\leq \alpha \Bigg(\sum^{L}_{i=1} \sup_{\bm{z}^{i-1}, \bm{\theta}^i} \Vert \bm{h}^i(\bm{z}^{i-1}, \bm{\theta}^{i}) - \widetilde{\bm{h}}^i(\bm{z}^{i-1}, \bm{\theta}^{i}) \Vert_2
+ \sqrt{\sum^{L}_{i=1} \sup_{\bm{z}^{i-1}, \bm{\theta}^i} \Vert \bm{h}^i(\bm{z}^{i-1}, \bm{\theta}^{i}) - \widetilde{\bm{h}}^i(\bm{z}^{i-1}, \bm{\theta}^{i}) \Vert_2} \ \Bigg).
\end{equation}
\end{proof}

\begin{proof}[Proof of \cref{thm:abpt}]
In Approx-BP training, an update step of parameters is denoted by
\begin{equation}
\bm{\theta}_{t+1} = \bm{\theta}_{t} - \eta \widehat{\bm{g}}_t.
\end{equation}
By Assumption 4.1 in \cite{bottou2018optimization}, we have
\begin{equation}
\begin{aligned}
\ell(\bm{f}(\bm{x}_t, \bm{\theta}_{t+1})) \leq &\ell(\bm{f}(\bm{x}_t, \bm{\theta}_{t})) + \bm{g}_t^\top (\bm{\theta}_{t+1} - \bm{\theta}_t) + \frac{\beta}{2} \Vert \bm{\theta}_{t+1} - \bm{\theta}_t \Vert_2^2 \\
= &\ell(\bm{f}(\bm{x}_t, \bm{\theta}_{t})) - \eta \bm{g}_t^\top \hat{\bm{g}}_t + \frac{\eta^2\beta}{2} \Vert \hat{\bm{g}}_t \Vert_2^2.
\end{aligned}
\end{equation}
Then, using assumption $\eta < \frac{1}{2\beta}$, we have
\begin{equation}
\begin{aligned}
\ell(\bm{f}(\bm{x}_t, \bm{\theta}_{t+1})) - \ell(\bm{f}(\bm{x}_t, \bm{\theta}_{t})) \leq &- \eta \bm{g}_t^\top (\bm{g}_t - \bm{g}_t + \hat{\bm{g}}_t) + \frac{\eta^2\beta}{2} \Vert \bm{g}_t - \bm{g}_t + \hat{\bm{g}}_t \Vert_2^2 \\
\leq &-\eta \Vert \bm{g}_t \Vert_2^2 + \eta \Vert \bm{g}_t - \hat{\bm{g}}_t \Vert \Vert \bm{g}_t \Vert + \eta^2\beta \Vert \bm{g}_t \Vert_2^2 + \eta^2 \beta \Vert \bm{g}_t - \hat{\bm{g}}_t \Vert_2^2 \\
\leq &-\frac{\eta}{2}(\Vert \bm{g}_t \Vert - \Vert \bm{g}_t - \hat{\bm{g}}_t \Vert)^2 + \eta \Vert \bm{g}_t - \hat{\bm{g}}_t \Vert_2^2 \\
\leq &-\frac{\eta}{4}\Vert \bm{g}_t \Vert_2^2 + \frac{3 \eta}{2} \Vert \bm{g}_t - \hat{\bm{g}}_t \Vert_2^2.
\end{aligned}
\end{equation}
Now we obtain that
\begin{equation}
\Vert \bm{g}_t \Vert_2^2 \leq \frac{4}{\eta}(\ell(\bm{f}(\bm{x}_t, \bm{\theta}_t)) - \ell(\bm{f}(\bm{x}_t, \bm{\theta}_{t+1}))) + 6 \Vert \bm{g}_t - \hat{\bm{g}}_t \Vert_2^2.
\label{eq:25}
\end{equation}
Taking expectation of \eqref{eq:25} for $\bm{x}_t\sim\mathcal{D}$, we have
\begin{equation}
\label{eq:26}
\mathbb{E}_\mathcal{D}\Vert \nabla_{\bm{\theta}} \ell (\bm{f}(\bm{x}, \bm{\theta}_t)) \Vert_2^2 \leq \frac{4}{\eta} [\mathbb{E}_\mathcal{D}\ell(\bm{f}(\bm{x}, \bm{\theta}_t)) - \mathbb{E}_\mathcal{D}\ell(\bm{f}(\bm{x}, \bm{\theta}_{t+1}))] + 6\sigma^2.
\end{equation}
Taking average of \eqref{eq:26} over $t = 0, ..., T-1$, we have
\begin{equation}
\begin{aligned}
\frac{1}{T} \mathop{\sum}_{t=0}^{T-1} \mathbb{E}_\mathcal{D} \Vert \nabla_{\bm{\theta}} \ell(\bm{f}(\bm{x}, \bm{\theta}_t)) \Vert_2^2
\leq \frac{4}{\eta T} [\mathbb{E}_\mathcal{D}\ell(\bm{f}(\bm{x}, \bm{\theta}_0)) - \mathbb{E}_\mathcal{D}\ell(\bm{f}(\bm{x}, \bm{\theta}_T))] + 6\sigma^2
\leq \frac{4}{\eta T} [\mathbb{E}_\mathcal{D}\ell(\bm{f}(\bm{x}, \bm{\theta}_0)) - \ell^*] + 6\sigma^2.
\end{aligned}
\end{equation}
Therefore, we conclude that
\begin{equation}
\begin{aligned}
\mathop{\min}_{t \in \{0, ..., T-1\}} &\mathbb{E}_\mathcal{D} \Vert \nabla_{\bm{\theta}} \ell(\bm{f}(\bm{x},\bm{\theta}_t)) \Vert^2_2
\leq \frac{4[\mathbb{E}_\mathcal{D} \ell(\bm{f}(\bm{x}, \bm{\theta}_0)) - \ell^*]}{\eta T} + 6\sigma^2.
\end{aligned}
\end{equation}
\end{proof}

\section{Derivation of Our ReGELU2 and ReSiLU2}
\label{appendix:regelu2}
\subsection{Proposed ReGELU2}
We denote GELU as $h$, and
\begin{equation}
h(x) = \frac{x}{2}(1 + \mathrm{erf}(\frac{x}{\sqrt{2}})).
\end{equation}
Then we define the approximate activation function $\widetilde{h}_{\bm{a}, \bm{c}}$ of GELU $h$ as 
follows:
\begin{equation}
\widetilde{h}_{\bm{a}, \bm{c}} (x) = a_1 \mathrm{max}\{x - c_1, 0\} + a_2 \mathrm{max}\{x - c_2, 0\} + (1 - a_1 - a_2) \mathrm{max}\{x - c_3, 0\}.
\end{equation}
The optimization objective is
\begin{equation}
\mathop{\mathrm{min}}_{\bm{a},\bm{c}} \int_{-\infty}^{\infty} (h(x) - \widetilde{h}_{\bm{a}, \bm{c}}(x))^2 \mathrm{d}x.
\end{equation}
We first perform a tail estimation for the integral in the objective.
Note that $\widetilde{h}_{\bm{a},\bm{c}}(x) \equiv 0$ for $x < \mathrm{min}\{\bm{c}\}$, \ie, the minimal value in the vector $\bm{c}$, and $\widetilde{h}_{\bm{a},\bm{c}}(x) \equiv x$ for $x > \mathrm{max}\{\bm{c}\}$, \ie, the maximum value in the vector $\bm{c}$.
So the left tail of the integral can be estimated as follows, for a certain $A < 0$:
\begin{equation}
\begin{aligned}
&\int_{-\infty}^A(h(x) - \widetilde{h}_{\bm{a},\bm{c}}(x))^2 \mathrm{d}x = \int_{-\infty}^A(\frac{x}{2}(1 + \mathrm{erf}(\frac{x}{\sqrt{2}})))^2 \mathrm{d}x \\
< &\int_{-\infty}^A - \frac{x}{2}(1 + \mathrm{erf}(\frac{x}{\sqrt{2}})) \mathrm{d}x = \int_{-\infty}^{\frac{A}{\sqrt{2}}} - x(1 + \mathrm{erf}(x)) \mathrm{d}x \\
< &\int_{-\infty}^{\frac{A}{\sqrt{2}}} - x(1 - \sqrt{1 - e^{-x^2}}) \mathrm{d}x < \int_{-\infty}^{\frac{A}{\sqrt{2}}} - x e^{-x^2} \mathrm{d}x = \frac{1}{2}e^{-\frac{A^2}{2}}.
\end{aligned}
\end{equation}

The right tail of the integral can be estimated as follows, for a certain $B > 0$:
\begin{equation}
\begin{aligned}
&\int_{B}^{+\infty}(h(x) - \widetilde{h}_{\bm{a},\bm{c}}(x))^2 \mathrm{d}x = \int_{B}^{+\infty}(\frac{x}{2}(1 - \mathrm{erf}(\frac{x}{\sqrt{2}})))^2 \mathrm{d}x \\
< &\int_{B}^{+\infty} \frac{x}{2}(1 - \mathrm{erf}(\frac{x}{\sqrt{2}})) \mathrm{d}x = \int_{\frac{B}{\sqrt{2}}}^{+\infty} x(1 - \mathrm{erf}(x)) \mathrm{d}x \\
< &\int_{\frac{B}{\sqrt{2}}}^{+\infty} x(1 - \sqrt{1 - e^{-x^2}}) \mathrm{d}x < \int_{\frac{B}{\sqrt{2}}}^{+\infty} x e^{-x^2} \mathrm{d}x = \frac{1}{2}e^{-\frac{B^2}{2}}.
\end{aligned}
\end{equation}

The condition of scaling inequalities above can be summarized as $\frac{|x|}{2}(1 - \mathrm{erf}(\frac{|x|}{\sqrt{2}})) < 1$ for $|x| > \mathrm{max}\{|A|, |B|\}$.
When setting $B = -A = \sqrt{-2 \mathrm{ln}(\varepsilon)}$, we have the following bounds:
\begin{equation}
\int_{-\infty}^A (h(x) - \widetilde{h}_{\bm{a},\bm{c}}(x))^2 \mathrm{d}x + \int_B^{+\infty} (h(x) - \widetilde{h}_{\bm{a},\bm{c}}(x))^2 \mathrm{d}x < \varepsilon.
\end{equation}
We set $\varepsilon = 10^{-8}$ to satisfy the condition of scaling inequalities and bound the two-side tails of integral in a negligible value.

Now we only need to solve the following optimization objective:
\begin{equation}
\mathop{\mathrm{min}}_{\bm{a},\bm{c}} \int_{A}^{B} (h(x) - \widetilde{h}_{\bm{a},\bm{c}}(x))^2 \mathrm{d}x.
\end{equation}
This time, the integral in the objective is a definite integral over a bounded interval, which can be calculated by many numerical  computing methods~\cite{piessens1983quadpack,2020SciPy-NMeth}.
Although the above optimization objective is not convex, it is not difficult to find a good solution, since there are only five 
scalar variables.
We have tried simulated annealing algorithm~\cite{kirkpatrick1983optimization} and stochastic gradient descent algorithm~\cite{robbins1951stochastic}, and both can find good solutions that are close to each other, as long as searching multiple times with different initialization.
The following solution is obtained by simulated annealing algorithm~\cite{kirkpatrick1983optimization}, which is adopted in our code:
\begin{equation*}
\begin{aligned}
\bm{a}^* &= [-0.04922261145617846, 1.0979632065417297]^\top,\\
\bm{c}^* &= [-3.1858810036855245, -0.001178821281161997, 3.190832613414926]^\top.
\end{aligned}
\end{equation*}

\begin{figure}[h]
\centering
\begin{minipage}{0.49\linewidth}
\centering
\includegraphics[width=\columnwidth]{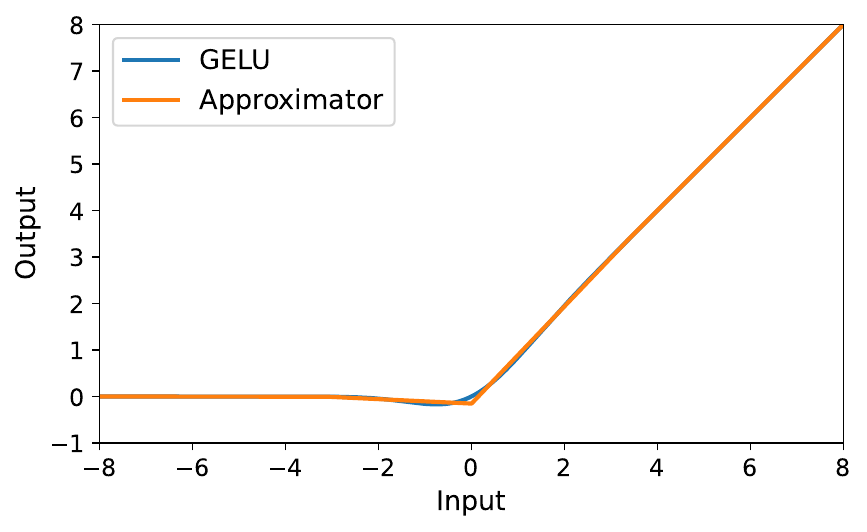}
\end{minipage}
\begin{minipage}{0.49\linewidth}
\centering
\includegraphics[width=\columnwidth]{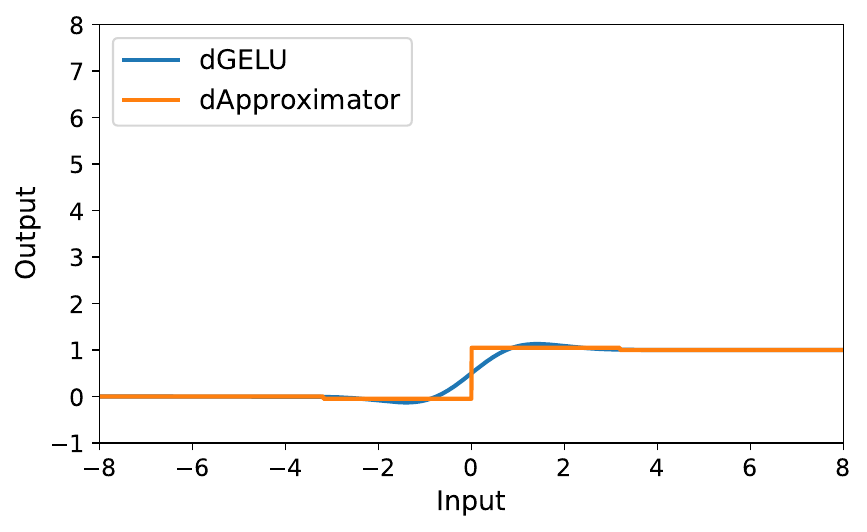}
\end{minipage}
\caption{\textbf{Plot curve of our ReGELU2}. The primitive function is the same as GELU. The derivative function is the same as the dApproximator (derivative of the approximate activation function $\widetilde{h}_{\bm{a}^*, \bm{c}^*}$ of GELU $h$), a 4-segment step function that needs 2
bits to store the derivative information of each element.}
\label{fig:regelu2_2}
\end{figure}

We plot our ReGELU2 in \cref{fig:regelu2_2}.
In principle, there should be an additional operation during or after the optimization to compel the solutions to fulfill the constraint in \eqref{eq:relus_family}.
However, we found the constraint is already satisfied due to the inherent property of the L2 metrics.

\subsection{Proposed ReSiLU2}
The derivation of our ReSiLU2 is similar to that for our ReGELU2.
We also denote SiLU as $h$,
\begin{equation}
h(x) = \frac{x}{1 + e^{-x}}.
\end{equation}
And our optimization objective is the same as ReGELU2,
\begin{equation}
\mathop{\mathrm{min}}_{\bm{a},\bm{c}} \int_{-\infty}^{\infty} (h(x) - \widetilde{h}_{\bm{a},\bm{c}}(x))^2 \mathrm{d}x.
\end{equation}
Again, we perform a tail estimation for the integral in the objective.
Since $\widetilde{h}_{\bm{a},\bm{c}}(x) \equiv 0$ for $x < \mathrm{min}\{\bm{c}\}$, \ie, the minimal value in the vector $\bm{c}$, and $\widetilde{h}_{\bm{a},\bm{c}}(x) \equiv x$ for $x > \mathrm{max}\{\bm{c}\}$, \ie, the maximum value in the vector $\bm{c}$, the left tail of the integral can be estimated as follows, for a certain $A < 0$:
\begin{equation}
\begin{aligned}
&\int_{-\infty}^A(h(x) - \widetilde{h}_{\bm{a},\bm{c}}(x))^2 \mathrm{d}x
= \int_{-\infty}^A(\frac{x}{1 + e^{-x}})^2 \mathrm{d}x \\
< &\int_{-\infty}^A\frac{-x}{1 + e^{-x}} \mathrm{d}x
< \int_{-\infty}^A - x e^x \mathrm{d}x
= (1 - A) e^A
< e^{\frac{A}{2}}.
\end{aligned}
\end{equation}

The right tail of the integral can be estimated as follows, for a certain $B < 0$:
\begin{equation}
\begin{aligned}
&\int_{B}^{+\infty}(h(x) - \widetilde{h}_{\bm{a},\bm{c}}(x))^2 \mathrm{d}x
= \int_{B}^{+\infty} (\frac{x}{1 + e^x})^2 \mathrm{d}x \\
< &\int_{B}^{+\infty} \frac{x}{1 + e^x} \mathrm{d}x
< \int_{B}^{+\infty} x e^{-x} \mathrm{d}x
= (1 + B) e^{-B} < e^{-\frac{B}{2}}.
\end{aligned}
\end{equation}

The condition of scaling inequalities above can be summarized as $\frac{|x|}{1 + e^{|x|}} < 1$ for $|x| > \mathrm{max}\{|A|, |B|\}$ and $1 - A < e^{-\frac{A}{2}}$ and $1 + B < e^{\frac{B}{2}}$.
When setting $B = -A = -2\mathrm{ln}(\frac{\varepsilon}{2})$, we have the following bounds:
\begin{equation}
\int_{-\infty}^A (h(x) - \widetilde{h}_{\bm{a},\bm{c}}(x))^2 \mathrm{d}x + \int_B^{+\infty} (h(x) - \widetilde{h}_{\bm{a},\bm{c}}(x))^2 \mathrm{d}x < \varepsilon.
\end{equation}
We set $\varepsilon = 10^{-8}$ to satisfy the condition of scaling inequalities and bound the two-side tails of integral in a negligible value.
Now we only need to consider the following optimization objective:
\begin{equation}
\mathop{\mathrm{min}}_{\bm{a},\bm{c}} \int_{A}^{B} (h(x) - \widetilde{h}_{\bm{a},\bm{c}}(x))^2 \mathrm{d}x.
\end{equation}
This time, the integral in the objective is a definite integral over a bounded interval, which can also be calculated be many numerical methods~\cite{piessens1983quadpack,2020SciPy-NMeth}.
Similarly, although the above optimization objective is not convex, it is not difficult to find a good solution, since there are only five 
scalar variables.
We have tried simulated annealing algorithm~\cite{kirkpatrick1983optimization} and stochastic gradient descent algorithm~\cite{robbins1951stochastic}, and both can find good solutions that are close to each other, as long as searching multiple times with different initialization.
The following solution is obtained by simulated annealing algorithm~\cite{kirkpatrick1983optimization}, which is adopted in our code:
\begin{equation*}
\begin{aligned}
\bm{a}^* &= [-0.04060357190528599, 1.080925428529668]^\top,\\
\bm{c}^* &= [-6.3050461001646445, -0.0008684942046214787, 6.325815242089708]^\top.
\end{aligned}
\end{equation*}

\begin{figure}[h]
\centering
\begin{minipage}{0.49\linewidth}
\centering
\includegraphics[width=\columnwidth]{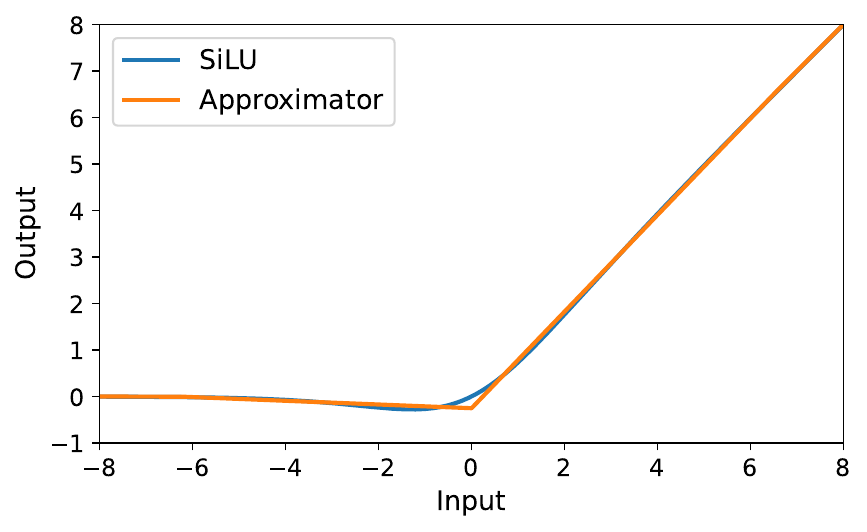}
\end{minipage}
\begin{minipage}{0.49\linewidth}
\centering
\includegraphics[width=\columnwidth]{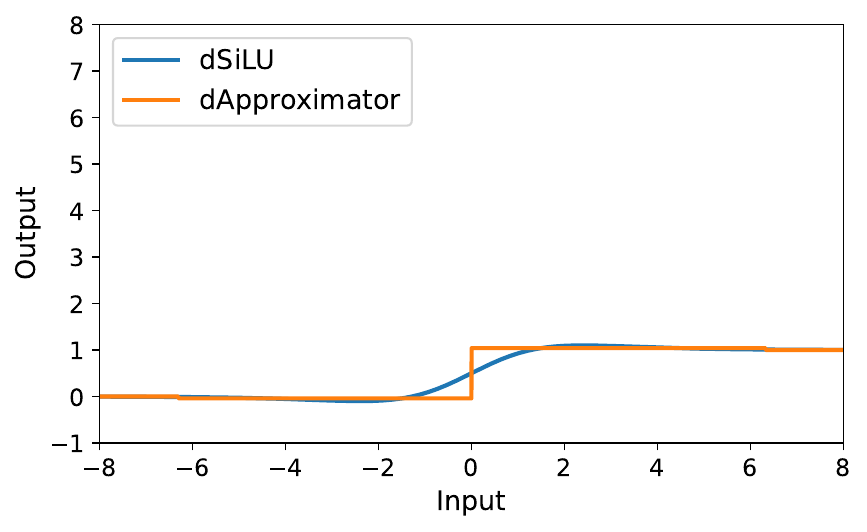}
\end{minipage}
\caption{\textbf{Plot curve of our ReSiLU2}. The primitive function is the same as SiLU. The derivative function is the same as the dApproximator (derivative of the approximate activation function $\widetilde{h}_{\bm{a}^*, \bm{c}^*}$ of SiLU $h$), a 4-segment step function that needs 2
bits to store the derivative information of each element.}
\label{fig:resilu2}
\end{figure}

We plot our ReSiLU2 in \cref{fig:resilu2}.
In principle, there should be an additional operation during or after the optimization to compel the solutions to fulfill the constraint in \eqref{eq:relus_family}.
However, we found the constraint is already satisfied due to the inherent property of the L2 metrics.

% \begin{figure}[h]
% \begin{center}
% \centerline{\includegraphics[width=\columnwidth/2]{resilu2}}
% \vskip -0.2in
% \caption{The line plot ReGELU2. The primitive function is the same as GELU. The derivative function is a 4-segment step function that need 2
% bits to store the derivative information.}
% \label{fig:resilu2}
% \end{center}
% \vskip -0.2in
% \end{figure}

\section{Memory-Sharing Activation Function}
\label{appendix:ms_act}
Suppose $\bm{h}^i$ is a layer of element-wise activation function.
The forward pass at $\bm{h}^i$ can be expressed as:
\begin{equation}
\bm{z}^i = \bm{h}^i(\bm{z}^{i-1}).
\end{equation}
The backward pass at $\bm{h}^i$ can be expressed as:
\begin{equation}
\frac{\partial \ell}{\partial \bm{z}^{i-1}} = \frac{\partial \bm{h}^i(\bm{z}^{i-1})}{\partial \bm{z}^{i-1}} \frac{\partial \ell}{\partial \bm{z}^{i}}.
\end{equation}
The first condition of \cref{proposition:ms_bp} is immediately satisfied.
The third condition of \cref{proposition:ms_bp} depends on the model architecture and the fine-tuning methods.
Here, we mainly consider the second condition of \cref{proposition:ms_bp}.
Since $\bm{h}^i$ is element-wise, we denote the scalar activation function in $\bm{h}^i$ as $h$.
Now, the second condition of \cref{proposition:ms_bp} can be rephrased as $\mathrm{d}h(x) = J(h(x))$, where $J$ is a certain function.
Some simple activation functions, such as ReLU and Sigmoid, satisfy this condition apparently:
\begin{equation}
\begin{aligned}
\mathrm{d}\mathrm{ReLU}(x) &= \mathrm{sgn}(\mathrm{ReLU}(x)),
\\
\mathrm{d}\sigma(x) &= \sigma(x)(1 - \sigma(x)),
\end{aligned}
\end{equation}
where ``$\text{sgn}$'' is the sign function and $\sigma(x)$ is the Sigmoid function.

However, it is challenging to answer whether a complicated activation function like SiLU satisfies this condition.
Here, we conclude that SiLU does not satisfy such condition.
To show this, we first give the analytic form of $h(x)$ and $\mathrm{d}h(x)$:
\begin{equation}
\begin{aligned}
h(x) &= x \sigma(x), \\
\mathrm{d}h(x) &= \sigma(x) + x \sigma(x) - x \sigma(x)^2 \\
&= \frac{h(x)- h(x)^2}{x} + h(x).
\end{aligned}
\label{eq:silu_dsilu}
\end{equation}
If $\mathrm{d}h(x) = J(h(x))$ for some function $J$, then $\mathrm{d}h(x)$ is decided only by $h(x)$.
Since $h(x)$ is not injective, there exits $x_1 \neq x_2$ such that $h(x_1) = h(x_2) \notin \{0, 1\}$, which derive $\mathrm{d}h(x_1) - \mathrm{d}h(x_2) = J(h(x_1)) - J(h(x_2)) = 0$.
However, from \eqref{eq:silu_dsilu}, we also derive  $\mathrm{d}h(x_1) - \mathrm{d}h(x_2) = (h(x_1)- h(x_1)^2)(\frac{1}{x_1} - \frac{1}{x_2}) \neq 0$, resulting in a contradiction.

\section{Memory-Sharing LayerNorm and RMSNorm}
\label{appendix:merge_rmsn}
\subsection{Proposed Memory-Sharing LayerNorm (MS-LN)}
The forward pass at LayerNorm and its following linear layer is as follows:
\begin{equation}
\begin{aligned}
& \text{Suppose} \ \bm{z}^{i-1} \in \mathbb{R}^{p_{i-1}}, \bm{H} = \mathbbm{I} - p_{i-1}^{-1}\mathbbm{1}\mathbbm{1}^\top,\\
&\sigma =  \sqrt{p_{i-1}^{-1} {\bm{z}^{i-1}}^\top \bm{H} \bm{z}^{i-1} + \varepsilon},\\
&\widetilde{\bm{z}}^{i-1} =  \sigma^{-1} \bm{H} \bm{z}^{i-1},\\
&\bm{z}^{i} = {\rm diag}(\bm{\alpha}) \widetilde{\bm{z}}^{i-1} + \bm{\beta},\\
&\bm{z}^{i+1} = \bm{W}\bm{z}^{i} + \bm{b}.
\end{aligned}
\end{equation}
We can merge the affine parameters in LayerNorm and the parameters in the following linear layer as follows:
\begin{equation}
\begin{aligned}
&\widetilde{\bm{W}} = \bm{W}{\rm diag}(\bm{\alpha}),\\
&\widetilde{\bm{b}} = \bm{W}\bm{\beta} + \bm{b}.
\end{aligned}
\end{equation}
Then the forward pass at a merged LayerNorm and the following linear layer becomes:
\begin{equation}
\begin{aligned}
& \text{Suppose} \ \bm{x} \in \mathbb{R}^{p_{i-1}}, \bm{H} = \mathbbm{I} - p_{i-1}^{-1}\mathbbm{1}\mathbbm{1}^\top,\\
&\sigma =  \sqrt{p_{i-1}^{-1} {\bm{z}^{i-1}}^\top \bm{H} \bm{z}^{i-1} + \varepsilon},\\
&\bm{z}^i = \sigma^{-1} \bm{H}\bm{z}^{i-1},\\
&\bm{z}^{i+1} = \widetilde{\bm{W}} \bm{z}^i + \widetilde{\bm{b}}.
\end{aligned}
\end{equation}
The program of our MS-LN is shown in \cref{alg:MS-LN}.

\begin{minipage}{.49\textwidth}
\begin{algorithm}[H]
   \caption{Memory-Sharing LayerNorm (MS-LN)}
   \label{alg:MS-LN}
\begin{algorithmic}
    \STATE Suppose $\bm{H} = \mathbbm{I} - p_{i-1}^{-1}\mathbbm{1}\mathbbm{1}^\top$, $\ell$ is the loss function.
    \STATE {\bfseries Input:} $\bm{z}^{i-1} \in \mathbb{R}^{p_{i-1}}$
    \STATE {\bfseries Forward:}
    \STATE \quad $\sigma = \sqrt{p_{i-1}^{-1} {\bm{z}^{i-1}}^\top \bm{H} \bm{z}^{i-1} + \varepsilon}$
    \STATE \quad $\bm{z}^{i} =  \sigma^{-1} \bm{H}\bm{z}^{i-1}$
    \STATE \quad Save for backward: $\bm{z}^{i}$, $\sigma$
    \STATE \quad {\bfseries Return Output:} $\bm{z}^{i}$
    \STATE {\bfseries Backward:}
    \STATE \quad Receive gradient: $\frac{\partial \ell}{\partial \bm{z}^{i}}$
    \STATE \quad $\frac{\partial \ell}{\partial \bm{z}^{i-1}} = \sigma^{-1} (\bm{H} - p_{i-1}^{-1} \bm{z}^{i} {\bm{z}^{i}}^\top) \frac{\partial \ell}{\partial \bm{z}^{i}}$
    \STATE \quad {\bfseries Return Gradient:} $\frac{\partial \ell}{\partial \bm{z}^{i-1}}$
\end{algorithmic}
\end{algorithm}
\end{minipage}\hfill
\begin{minipage}{.49\textwidth}
\begin{algorithm}[H]
   \caption{Memory-Sharing RMSNorm (MS-RMSNorm)}
   \label{alg:MS-RMSNorm}
\begin{algorithmic}
    \STATE Suppose $\ell$ is the loss function.
    \STATE {\bfseries Input:} $\bm{z}^{i-1} \in \mathbb{R}^{p_{i-1}}$
    \STATE {\bfseries Forward:}
    \STATE \quad $\sigma = \sqrt{p_{i-1}^{-1} {\bm{z}^{i-1}}^\top \bm{z}^{i-1} + \varepsilon}$
    \STATE \quad $\bm{z}^{i} = \sigma^{-1} \bm{z}^{i-1}$
    \STATE \quad Save for backward: $\bm{z}^{i}$, $\sigma$
    \STATE \quad {\bfseries Return Output:} $\bm{z}^i$
    \STATE {\bfseries Backward:}
    \STATE \quad Receive gradient: $\frac{\partial \ell}{\partial \bm{z}^i}$
    \STATE \quad $\frac{\partial \ell}{\partial \bm{z}^{i-1}} = \sigma^{-1} (\mathbb{I} - p_{i-1}^{-1} \bm{z}^i {\bm{z}^i}^\top) \frac{\partial \ell}{\partial \bm{z}^i}$
    \STATE \quad {\bfseries Return Gradient:} $\frac{\partial \ell}{\partial \bm{z}^{i-1}}$
\end{algorithmic}
\end{algorithm}
\end{minipage}

\subsection{Proposed Memory-Sharing RMSNorm (MS-RMSNorm)}
The forward pass at RMSNorm and its following linear layer is as follows:
\begin{equation}
\begin{aligned}
& \text{Suppose} \ \bm{z}^{i-1} \in \mathbb{R}^{p_{i-1}},\\
&\sigma =  \sqrt{p_{i-1}^{-1} {\bm{z}^{i-1}}^\top \bm{z}^{i-1} + \varepsilon},\\
&\widetilde{\bm{z}}^{i-1} =  \sigma^{-1} \bm{z}^{i-1},\\
&\bm{z}^i = {\rm diag}(\bm{\alpha}) \widetilde{\bm{z}}^{i-1},\\
&\bm{z}^{i+1} = \bm{W}\bm{z}^i + \bm{b}.
\end{aligned}
\end{equation}
We can merge the affine parameters in RMSNorm and the parameters in the following linear layer as follows:
\begin{equation}
\begin{aligned}
\widetilde{\bm{W}} = \bm{W}{\rm diag}(\bm{\alpha}).
\end{aligned}
\end{equation}
Then the forward pass at a merged RMSNorm and the following linear layer becomes as follows:
\begin{equation}
\begin{aligned}
& \text{Suppose} \ \bm{z}^{i-1} \in \mathbb{R}^{p_{i-1}},\\
&\sigma =  \sqrt{p_{i-1}^{-1} {\bm{z}^{i-1}}^\top \bm{z}^{i-1} + \varepsilon},\\
&\bm{z}^{i} =  \sigma^{-1} \bm{z}^{i-1},\\
&\bm{z}^{i+1} = \widetilde{\bm{W}}\bm{z}^{i} + \bm{b}.
\end{aligned}
\end{equation}
The program of our MS-RMSNorm is shown in \cref{alg:MS-RMSNorm}.

\section{Implementation Details of Fine-Tuning ViT, LLaMA and RoBERTa in Our Experiments}
\label{appendix:implementation_details}
For experiments on fine-tuning ViT-base and ViT-large with LoRA and LoRA-FA, we use slight data augmentations in our experiments, which are Resize (to 224$\times$224 px), RandomCrop, RandomHorizontalFlip, Normalize for the train set and Resize (to 224$\times$224 px), CenterCrop, Normalize for the test set.
We use AdamW \cite{loshchilov2017decoupled} with the weight decay $0.1$ in all our experiments on ViTs.
The batch size is set as 64.
All ViT models are fine-tuned with WarmUp in the first 10 epochs, where the initial learning rate starts from $1$e-6, and Cosine learning rate scheduler in the remaining 90 epochs.
The base learning rate is $1.25$e-3 in LoRA and $1.25$e-5 in Full Tuning.
ViT-base experiments are conducted with 1$\times$2080Ti GPU and ViT-large experiments are conducted with 1$\times$L40 GPU.
We use automatic mixed precision (AMP) in \rm{Pytorch} as the default setting.
For experiments on fine-tuning LLaMA-7B and LLaMA-13B with QLoRA, the batch size is set as $4$ and the number of gradient accumulation steps is set as $4$.
The total training iterations are $10000$ steps.
For LLaMA-7B, we use paged AdamW with no weight decay, tune constant learning rate in $\{10^{-4}, 2\times10^{-4}\}$, and report the best 5-shot MMLU accuracy among them.
For LLaMA-13B, we tune learning rate in $\{10^{-4}, 2\times10^{-4}\}$, while setting weight decay as $0$ for \{SiLU, RMSNorm\} and \{ReSiLU2, RMSNorm\} configurations.
We set learning rate as 1e-4, while tuning weight decay in $\{0.1, 0.2\}$ for \{SiLU, MS-RMSNorm\} and \{ReSiLU2, MS-RMSNorm\} configurations.
Gradient checkpointing~\cite{chen2016training} is not used in our experiments.
For experiments on fine-tuning RoBERTa-base with LoRA, the batch size is set as $32$.
We use AdamW with the weight decay $0.01$.
All RoBERTa-base models are fine-tuned from the pretrained model independently for $30$ epochs.
We use Linear learning rate scheduler with WarmUp ratio $0.1$.
The base learning rate for each task is chosen as the best one among $\{0.00005, 0.0001, 0.0005, 0.001, 0.005\}$ in fine-tuning the baseline.

\section{Choice of Optimization Objective for Approximate Activation Function $\widetilde{h}_{\bm{a},\bm{c}}(x)$}
\label{appendix:choices_of_optimization_objective}
In \cref{method:regelu2}, we derive the optimization objective~\eqref{eq:objective} from our Approx-BP theory.
Meanwhile, we believe that there exist other feasible choices of optimization objective.
A heuristic choice can be,
\begin{equation}
\mathop{\mathrm{min}}_{\bm{a},\bm{c}} \int_{-\infty}^{\infty} (\mathrm{d}h(x) - \mathrm{d}\widetilde{h}_{\bm{a},\bm{c}}(x))^2 \mathrm{d}x.
\label{eq:objective_d}
\end{equation}
Applying the similar technique introduced in \cref{appendix:regelu2} to the above optimization problem \eqref{eq:objective_d}, we obtain another alternative of GELU.
We call this alternative as ReGELU2-d, which means ReGELU2-d directly approximates the derivatives of GELU.
The according solution of \eqref{eq:objective_d} for $\{\bm{a},\bm{c}\}$ is:
\begin{equation*}
\begin{aligned}
\bm{a}^* &= [ 0.32465931184406527, 0.34812875668739607]^\top,\\
\bm{c}^* &= [-0.4535743722857079, -0.0010587205574873046, 0.4487575313884231]^\top.
\end{aligned}
\end{equation*}
In our experiments (\cref{appendix:tab:vit-b_lora}), the fine-tuning ViT-base using LoRA with the new alternative ReGELU-d is also stable, but the results by ReGELU2-d are consistently inferior to those by our ReGELU2. Therefore, we still employ ReGELU2 and ReSiLU2 in our main paper.

\setlength{\tabcolsep}{4.0pt}
\begin{table*}[ht]
% \vspace{-5mm}
\caption{
\textbf{Results of fine-tuning ViT-base using LoRA with different activation functions} on the CIFAR10 (C10), CIFAR100 (C100), and FGVC benchmarks.
We report the Top-1 accuracy (\%) results on each dataset and the mean Top-1 accuracy (\%) results on all seven datasets.
The best results are highlighted in \textbf{bold}.}
\centering
\begin{adjustbox}{max width =1.0\linewidth}
\begin{tabular}{cll|ccccccc|c}
\toprule
\textbf{Method} & Activation & Norm & C10 & C100 & CUB & NAB & Flower & Dogs & Cars & Mean \\
\midrule
\multirow{3}{*}{\shortstack{LoRA\\$r = 4$\\Q, V}}
& GELU & LN & \textbf{98.8} & \textbf{92.0} & 86.7 & \textbf{83.2} & \textbf{99.3} & 90.7 & \textbf{81.5} & \textbf{90.3}\\
& ReGELU2-d & LN & 98.7 & \textbf{92.0} & 86.8 & 82.9 & \textbf{99.3} & 90.8 & 81.1 & 90.2\\
& ReGELU2 & LN & \textbf{98.8} & \textbf{92.0} & \textbf{86.9} & 83.0 & \textbf{99.3} &\textbf{91.0} & 81.4 & \textbf{90.3}\\
\midrule
\multirow{3}{*}{\shortstack{LoRA\\$r = 4$\\All Linear}}
& GELU & LN & \textbf{98.9} & \textbf{93.0} & \textbf{87.3} & \textbf{83.0} & \textbf{99.2} & 90.7 & 82.9 & 90.7\\
& ReGELU2-d & LN & \textbf{98.9} & 92.7 & 87.2 & 82.6 & \textbf{99.2} & 91.0 & 82.6 & 90.6\\
& ReGELU2 & LN & \textbf{98.9} & 92.8 & \textbf{87.3} & \textbf{83.0} & \textbf{99.2} & \textbf{91.1} & \textbf{83.2} & \textbf{90.8}\\
\bottomrule
\end{tabular}
\end{adjustbox}
\label{appendix:tab:vit-b_lora}
% \vspace{-5mm}
\end{table*}

\section{More Experiments Results}
\label{appendix:more_results}
In this section, some experimental results are supplementary to the main text, while others provide more diverse evaluations of our method.

\subsection{Experiments on ViT}
\label{appendix:vit_experiments}
The results in \cref{appendix:tab:vit-b_lora_all} are supplementary to those in \cref{tab:vit-b_lora}.
Here, we report the results of replacing the activation function of the pretrained ViT-base with ReLU as a reference.
The training throughput of GELU, ReLU, and ReGELU2 is similar, while the training performance of ReLU is significantly inferior to other activation functions in the comparison.
When all linear layers are adapted by LoRA, the reduction of GPU memory usage during fine-tuning is similar between ReLU and ReGELU2.
When only the query and value projections are adapted, ReLU can not reduce the GPU memory usage, whereas ReGELU2 can reduce the GPU memory usage by $\sim$$19\%$.
That indicates that ReLU is probably implemented in Pytorch in a manner as we described in \cref{appendix:ms_act}.

\begin{table*}[ht]
\caption{
\textbf{Results of fine-tuning ViT-base using LoRA or LoRA-FA with different activation function and layer normalization} on the CIFAR10 (C10), CIFAR100 (C100), and FGVC benchmarks.
We report the Top-1 accuracy (\%) results on each dataset and the mean Top-1 accuracy (\%) results on all seven datasets.
The best results are highlighted in \textbf{bold}.
}
\centering
\begin{adjustbox}{max width =1.0\linewidth}
\begin{tabular}{cll|ccccccc|cll}
\toprule
& & & \multicolumn{7}{c}{Dataset} & \multicolumn{3}{|c}{Mean} \\
Method & Activation & Norm & C10 & C100 & CUB & NAB & Flower & Dogs & Cars & Top-1(\%) & Mem.(MiB) & Thr.(images/s) \\
\midrule
\multirow{8}{*}{\shortstack{LoRA\\$r = 4$\\Q, V}} & GELU & LN & \textbf{98.8} & \textbf{92.0} & 86.7 & \textbf{83.2} & \textbf{99.3} & 90.7 & \textbf{81.5} & \textbf{90.3} & 3827 & 288\\
 & ReLU & LN & 98.4 & 90.4 & 85.5 & 81.8 & 97.4 & 88.4 & 80.7 & 89.0 & 3828(+0\%) & \textbf{290}(+1\%)\\
 & Mesa-GELU & LN & \textbf{98.8} & \textbf{92.0} & 86.6 & 83.1 & \textbf{99.3} & 90.8 & 81.1 & \textbf{90.3} & 3453(-10\%) & 245(-15\%) \\
 & ReGELU2 & LN & \textbf{98.8} & \textbf{92.0} & \textbf{86.9} & 83.0 & \textbf{99.3} & \textbf{91.0} & 81.4 & \textbf{90.3} & \textbf{3087}(-19\%) & 289(+0\%) \\
\cmidrule{2-13}
 & GELU & Mesa-LN & \textbf{98.8} & 91.8 & 86.8 & \textbf{82.9} & \textbf{99.2} & 90.8 & 81.3 & 90.2 & \textbf{3249}(-15\%) & 257(-11\%)\\
 & GELU & MS-LN & \textbf{98.8} & \textbf{92.3} & \textbf{88.1} & 82.7 & \textbf{99.2} & \textbf{90.9} & \textbf{83.1} & \textbf{90.7} & 3441(-10\%) & \textbf{288}(+0\%)\\
\cmidrule{2-13}
 & Mesa-GELU & Mesa-LN & \textbf{98.8} & 92.1 & 86.7 & \textbf{83.0} & \textbf{99.3} & \textbf{90.9} & \textbf{82.1} & 90.4 & 2853(-25\%) & 226(-22\%)\\
 & ReGELU2 & MS-LN & \textbf{98.8} & \textbf{92.3} & \textbf{88.0} & 82.6 & 99.2 & 90.8 & \textbf{82.1} & \textbf{90.5} & \textbf{2717}(-29\%) & \textbf{290}(+1\%)\\
\midrule
\multirow{8}{*}{\shortstack{LoRA\\$r = 4$\\All Linear}} & GELU & LN & \textbf{98.9} & \textbf{93.0} & \textbf{87.3} & \textbf{83.0} & \textbf{99.2} & 90.7 & 82.9 & 90.7 & 5128 & 207\\
 & ReLU & LN & 98.8 & 92.0 & 86.0 & 81.8 & 97.0 & 89.0 & 82.1 & 89.5 & \textbf{4300}(-16\%) & \textbf{208}(+0\%)\\
 & Mesa-GELU & LN & \textbf{98.9} & 92.9 & \textbf{87.3} & 82.8 & 99.0 & \textbf{91.2} & \textbf{83.3} & \textbf{90.8} & 4721(-8\%) & 186(-10\%)\\
 & ReGELU2 & LN & \textbf{98.9} & 92.8 & \textbf{87.3} & \textbf{83.0} & \textbf{99.2} & 91.1 & 83.2 & \textbf{90.8} & 4380(-15\%) & 207(+0\%)\\
\cmidrule{2-13}
 & GELU & Mesa-LN & 99.0 & 92.8 & 87.4 & \textbf{83.0} & \textbf{99.3} & \textbf{90.8} & 83.2 & 90.8 & 4530(-12\%) & 189(-9\%)\\
 & GELU & MS-LN & \textbf{99.1} & \textbf{93.0} & \textbf{88.5} & 82.9 & 99.2 & \textbf{90.8} & \textbf{85.0} & \textbf{91.2} & \textbf{4316}(-16\%) & \textbf{207}(+0\%)\\
\cmidrule{2-13}
 & Mesa-GELU & Mesa-LN & 98.9 & 92.9 & 87.2 & 82.9 & 99.3 & \textbf{91.1} & 83.2 & 90.8 & 4209(-18\%) & 173(-17\%)\\
 & ReGELU2 & MS-LN & \textbf{99.0} & \textbf{93.1} & \textbf{88.0} & \textbf{83.1} & \textbf{99.4} & 90.9 & \textbf{84.8} & \textbf{91.2} & \textbf{3601}(-30\%) & \textbf{208}(+0\%)\\
\midrule
\multirow{4}{*}{\shortstack{LoRA-FA\\$r = 4$\\Q, V}} & GELU & LN & \textbf{98.4} & 91.7 & \textbf{88.5} & 82.8 & \textbf{99.1} & 91.8 & \textbf{77.6} & \textbf{90.0} & 3386 & 304\\
 & Mesa-GELU & LN & \textbf{98.4} & \textbf{91.9} & 88.1 & 82.8 & \textbf{99.1} & 91.6 & \textbf{77.6} & 89.9 & 3012(-11\%) & 261(-14\%) \\
 & Mesa-GELU & Mesa-LN & 98.3 & 91.4 & 88.2 & \textbf{83.0} & \textbf{99.1} & 91.7 & 77.5 & 89.9 & \textbf{2411}(-29\%) & 236(-22\%) \\
 & ReGELU2 & LN & \textbf{98.4} & 91.7 & 88.1 & 82.6 & \textbf{99.1} & \textbf{91.9} & 77.2 & 89.8 & 2597(-23\%) & \textbf{306}(+1\%) \\
\midrule
\multirow{4}{*}{\shortstack{LoRA-FA\\$r = 4$\\All Linear}} & GELU & LN & 98.6 & 91.5 & \textbf{88.1} & \textbf{83.1} & 99.2 & 91.8 & 79.3 & \textbf{90.2} & 3430 & 249\\
 & Mesa-GELU & LN & 98.6 & \textbf{91.8} & 88.0 & 82.9 & 99.2 & \textbf{91.9} & 79.0 & \textbf{90.2} & 3021(-12\%) & 218(-12\%)\\
 & Mesa-GELU & Mesa-LN & \textbf{98.7} & 91.6 & 87.8 & 82.7 & \textbf{99.3} & 91.8 & 79.2 & 90.1 & \textbf{2457}(-28\%) & 200(-20\%)\\
 & ReGELU2 & LN & 98.6 & 91.7 & 88.0 & 82.9 & 99.1 & 91.8 & \textbf{79.4} & \textbf{90.2} & 2717(-21\%) & \textbf{251}(+0\%)\\
\bottomrule
\end{tabular}
\end{adjustbox}
\label{appendix:tab:vit-b_lora_all}
\end{table*}

\subsection{Experiments on LLaMA}
\label{appendix:llama_experiments}
As a supplementary material to \cref{tab:llama-7b-13b}, we report the BoolQ, PIQA, HS, WG, ARC-e, ARC-c, and OBQA metrics on fine-tuned LLaMA-7B in \cref{appendix:tab:llama-7b}.
We observe that the released checkpoint by the authors of QLoRA does not achieve much better results than the pretrained (without fine-tuning) LLaMA checkpoint.
Thus, we speculate that these metrics in \cref{appendix:tab:llama-7b} are not suitable to serve as the evaluation metrics for fine-tuning LLaMA-7b on Alpaca dataset.
However, our method still gets comparable performance on these metrics to the baseline.

We also have evaluated the max affordable training sequence length of LLaMA-7B with QLoRA on single RTX4090, which is summarized as \cref{appendix:tab:llama-7b-length}.
Our method can increase the max affordable training sequence length by $\sim 46\%$.

\setlength{\tabcolsep}{4.0pt}
\begin{table*}[h!]
\vspace{-1mm}
\caption{\textbf{Supplementary results on fine-tuning LLaMA-7B using QLoRA on Alpaca}.
The metrics are evaluated by "lm-evaluation-harness" package \cite{eval-harness}.
The best results are highlighted in \textbf{bold}.
}
\centering
\begin{adjustbox}{max width =1.0\linewidth}
\begin{tabular}{cl|cccccccc}
\toprule
Method & Checkpoint & BoolQ & PIQA & SIQA & HellaSwag & WinoGrande & ARC-e & ARC-c & OBQA \\
\midrule
\multirow{4}{*}{\shortstack{QLoRA\\$r = 64$\\All Linear}}
& pretrained                  & 74.43          & 78.45          & 32.91          & 75.00          & \textbf{70.09} & \textbf{71.25} & 44.54          & 44.80 \\
& officially released         & 72.02          & 78.73          & 32.65          & \textbf{76.05} & 69.61          & 68.90          & 46.42          & 43.20 \\
& fine-tuned by us            & \textbf{74.50} & 78.02          & 33.06          & 75.93          & 67.48          & 68.94          & 46.33          & 45.00 \\
& with ReSiLU2 and MS-RMSNorm & 73.76          & \textbf{79.54} & \textbf{33.21} & 75.82          & 68.43          & 69.53          & \textbf{47.18} & \textbf{45.60} \\
\bottomrule
\end{tabular}
\end{adjustbox}
\label{appendix:tab:llama-7b}
\vspace{-1mm}
\end{table*}

\setlength{\tabcolsep}{4.0pt}
\begin{table*}[h!]
\vspace{-1mm}
\caption{\textbf{Max affordable sequence length on fine-tuning LLaMA-7B using QLoRA}.
Batch size is set as 1.
The best results are highlighted in \textbf{bold}.
}
\centering
\begin{adjustbox}{max width =1.0\linewidth}
\begin{tabular}{cll|l}
\toprule
Method & Activation & Norm & Max Length of Tokens \\
\midrule
\multirow{4}{*}{\shortstack{QLoRA\\$r = 64$\\All Linear}}
& SiLU    & RMSNorm    & 1354 \\
& ReSiLU2 & RMSNorm    & 1504(+11\%) \\
& SiLU    & MS-RMSNorm & 1654(+22\%) \\
& ReSiLU2 & MS-RMSNorm & \textbf{1979}(+46\%) \\
\bottomrule
\end{tabular}
\end{adjustbox}
\label{appendix:tab:llama-7b-length}
\vspace{-1mm}
\end{table*}

\subsection{Experiments on SwinTransformer}
\label{appendix:swin_transformer_experiments}

We fine-tune the pretrained SwinTransformer-Tiny (Swin-T) and SwinTransformer-Small (Swin-S) \cite{liu2021Swin} with the detection head RetinaNet \cite{lin2017focal} on the PASCAL VOC object detection benchmark \cite{everingham2015the}.
This experiment is conducted by data parallel training using 4$\times$RTX2080Ti.
The reported peak memory usage is the max value of those from the 4 GPUs.
We use the training sets from VOC2007 and VOC2012 as the training set and the test set from VOC2007 as the test set.
The number of training epochs is set as 12.
The data type in this experiment is fp32.
The results are summarized in \cref{appendix:tab:swin_det}.
One can see that our method reduces $\sim18\%$ of the total memory consumption on fine-tuning Swin-T and Swin-S.

\setlength{\tabcolsep}{4.0pt}
\begin{table*}[th]
\vspace{-1mm}
\caption{\textbf{Results of fine-tuning SwinTransformer-tiny (Swin-T) and SwinTransformer-small (Swin-S)} with the detection head RetinaNet on the PASCAL VOC object detection benchmark.
The best results are highlighted in \textbf{bold}.
}
\centering
\begin{adjustbox}{max width =1.0\linewidth}
\begin{tabular}{llcll|llll}
\toprule
Head & Backbone & Batch Size & Activation & Norm & Mem.(MiB) & Min/Epoch & mAP & AP50 \\
\midrule
\multirow{4}{*}{RetinaNet}
& \multirow{2}{*}{Swin-T} & \multirow{2}{*}{4} & GELU    & LN    & 7026                 & 29.7                & \textbf{79.37} & \textbf{79.40} \\
&                         &                    & ReGELU2 & MS-LN & \textbf{5756}(-18\%) & \textbf{29.2(-2\%)} & 79.20          & 79.20 \\
\cmidrule{2-9}
& \multirow{2}{*}{Swin-S} & \multirow{2}{*}{2} & GELU    & LN    & 5810                 & 52.2                & \textbf{80.78} & \textbf{80.80} \\
&                         &                    & ReGELU2 & MS-LN & \textbf{4773}(-18\%) & \textbf{50.5}(-3\%) & 80.45          & 80.40 \\
\bottomrule
\end{tabular}
\end{adjustbox}
\label{appendix:tab:swin_det}
\vspace{-1mm}
\end{table*}

\subsection{Experiments on BERT}
\label{appendix:bert_experiment}
We fine-tune pretrained Bert-base \cite{devlin2018bert} on Squad-v2 \cite{rajpurkar2018know} benchmark using data parallel training by 4$\times$RTX3060.
The number of training epochs is 2.
The data type in this experiment is fp32.
The results are summarized in \cref{appendix:tab:bert_base}.
Our method enables to increase the batch size by $20\%$.

It is worth noting that increasing batch size usually enables less communication times, and thus larger throughput, in the distributed training.
To demonstrate that, we fine-tune pretrained Bert-large on Squad-v2 under the ZeRO training framework \cite{rasley2020deepspeed,rajbhandari2020zero,rajbhandari2021zeroinfinity} using 4$\times$RTX3060.
As shown in \cref{appendix:tab:bert_large}, our method can increase the throughput by $\sim 26\%$.

\setlength{\tabcolsep}{4.0pt}
\begin{table*}[th]
\vspace{-1mm}
\caption{\textbf{Results of fine-tuning Bert-base} on Squad-v2 using 4$\times$RTX3060 GPUs.
We set the batch size to the max affordable size.
The batch size in the table is the batch size per GPU.
The best results are highlighted in \textbf{bold}.
}
\centering
\begin{adjustbox}{max width =1.0\linewidth}
\begin{tabular}{lllc|lll}
\toprule
Model & Activation & Norm & Batch Size & Thr.(samples/s) & EM & F1 \\
\midrule
\multirow{2}{*}{Bert-base}
& GELU    & LN    & 30 & 76                & 70.94 & 74.14 \\
& ReGELU2 & MS-LN & 36 & \textbf{78}(+3\%) & \textbf{71.36} & \textbf{74.63} \\
\bottomrule
\end{tabular}
\end{adjustbox}
\label{appendix:tab:bert_base}
\vspace{-1mm}
\end{table*}

\setlength{\tabcolsep}{4.0pt}
\begin{table*}[th]
\vspace{-1mm}
\caption{\textbf{Results of fine-tuning Bert-large} on Squad-v2 using 4$\times$RTX3060 GPUs.
We set the batch size to the max affordable size.
The batch size in the table is the batch size per GPU.
The best results are highlighted in \textbf{bold}.
}
\centering
\begin{adjustbox}{max width =1.0\linewidth}
\begin{tabular}{llllc|llll}
\toprule
Model & ZeRO & Activation & Norm & Batch Size & Thr.(samples/s) & Hour/Epoch & EM & F1 \\
\midrule
\multirow{2}{*}{Bert-large} & \multirow{2}{*}{Stage 3 + CPU offload}
& GELU    & LN     & 10     & 9.57                  & 3.83                 & \textbf{77.29} & \textbf{80.65} \\
&& ReGELU2 & MS-LN & 14     & \textbf{12.03}(+26\%) & \textbf{3.05}(-20\%) & 77.19 & 80.59 \\
\bottomrule
\end{tabular}
\end{adjustbox}
\label{appendix:tab:bert_large}
\vspace{-1mm}
\end{table*}

%%%%%%%%%%%%%%%%%%%%%%%%%%%%%%%%%%%%%%%%%%%%%%%%%%%%%%%%%%%%%%%%%%%%%%%%%%%%%%%
%%%%%%%%%%%%%%%%%%%%%%%%%%%%%%%%%%%%%%%%%%%%%%%%%%%%%%%%%%%%%%%%%%%%%%%%%%%%%%%

\end{document}